%% file: main-arxiv.tex
\documentclass{article} 
\usepackage{iclr2027_conference,times}

\input{math_commands.tex}

\usepackage{amssymb}
\usepackage{hyperref}
\usepackage{url}
\usepackage{multirow}
\usepackage{booktabs}
\usepackage{makecell}
\usepackage{xcolor}
\usepackage{pifont}
\usepackage[normalem]{ulem}
\usepackage{graphicx}
\usepackage{wrapfig}
\usepackage{algorithm}
\usepackage[noend]{algpseudocode}

\usepackage[table]{xcolor}
\usepackage{booktabs,tabularx}

\title{ReLMem: Learning Recurrent Memory for Longitudinal EHR Modeling}

\iclrfinalcopy
\let\arxivmaketitle\maketitle
\renewcommand{\maketitle}{\arxivmaketitle\lhead{Preprint}}

\author{Zijie Meng$^{1}$\thanks{Equal contribution.} \quad Xiwei Dai$^{1}$\footnotemark[1] \quad Yingying Zhang$^2$ \quad Jian Wu$^1$ \quad Xian Wu$^{2}$\thanks{Corresponding authors.}  \quad Zuozhu Liu$^{1}$\footnotemark[2] \\
$^1$ Zhejiang University \quad $^2$ Tencent Jarvis Lab
}

\begin{document}
\maketitle

\begin{abstract}
Longitudinal electronic health record (EHR) modeling requires integrating new visits with an expanding patient history. Yet the continual accumulation of clinical information imposes increasing computational and memory costs on large language models (LLMs) when they process and retain complete patient histories. A practical alternative is visit-wise recurrent compression, which incorporates each incoming visit into a compact, continually updated patient memory. However, under a fixed memory budget, successive updates must integrate new information without progressively losing critical historical evidence needed to subsequent tasks. To address this challenge, we introduce \textbf{Re}current \textbf{L}ongitudinal \textbf{Mem}ory (\textbf{ReLMem}), a framework that learns to maintain fixed-capacity patient memory for efficient downstream prediction with a frozen LLM. ReLMem equips this LLM with lightweight compression adapters to recurrently update the memory from its previous state and each incoming visit, without rereading earlier records. Specifically, we develop a multi-granularity optimization strategy to preserve task-relevant information throughout recurrent updates and support downstream prediction from the final memory. The intermediate supervision aligns attention outputs from compressed memory and the full history under identical queries, while prediction supervision minimizes cross-entropy with ground truth answers conditioned on the final memory. On EHR-based medication prediction, ReLMem approaches the F1 scores of full-history baseline while reducing average retained historical storage by 97.1\%. Under the same memory budget, it improves macro- and micro-F1 over the strongest compressed-memory baseline by 4.66 and 4.75 percentage points, respectively. These results highlight the value of learning recurrent patient memory for efficient longitudinal EHR modeling.
\end{abstract}

\section{Introduction}
As a key component of modern healthcare, electronic health records (EHRs) have been widely adopted in clinical practice: as of 2024, over 99\% of non-federal acute care hospitals and 91\% of office-based physicians in the United States had adopted certified EHR systems \citep{onc2026ehradoption}. By documenting patients' health conditions, treatments, and outcomes across clinical encounters, EHRs support a wide range of complex tasks, including clinical decision-making, patient monitoring, and biomedical research \citep{jensen2012mining,moor2023foundation}. For each patient, successive visits contribute new observations, diagnoses, procedures, and treatments to an evolving longitudinal clinical history, which provides context for understanding the patient's current condition and anticipating future health outcomes \citep{choi2016retain,kraljevic2024foresight,shmatko2025delphi}. Therefore, effectively modeling these longitudinal visit trajectories is key to realizing the potential of EHRs.

Recent advances in large language models (LLMs) have introduced a new paradigm for modeling such complex clinical information in EHRs, with applications in clinical information extraction, medical forecasting, and multi-step clinical question answering \citep{yang2022gatortron,kraljevic2024foresight2,shi2024ehragent}. However, applying LLMs directly to an expanding patient history incurs increasing computational and memory overhead. For models with dense self-attention, the computational cost of attention during context prefill scales quadratically with sequence length, while the size of the historical key–value (KV) cache grows linearly \citep{yang2025qwen3}. Although cache reuse avoids re-encoding previously processed records, processing each incoming visit still requires attention over the accumulated history.

\begin{figure}[t]
    \centering
    \includegraphics[width=\textwidth]{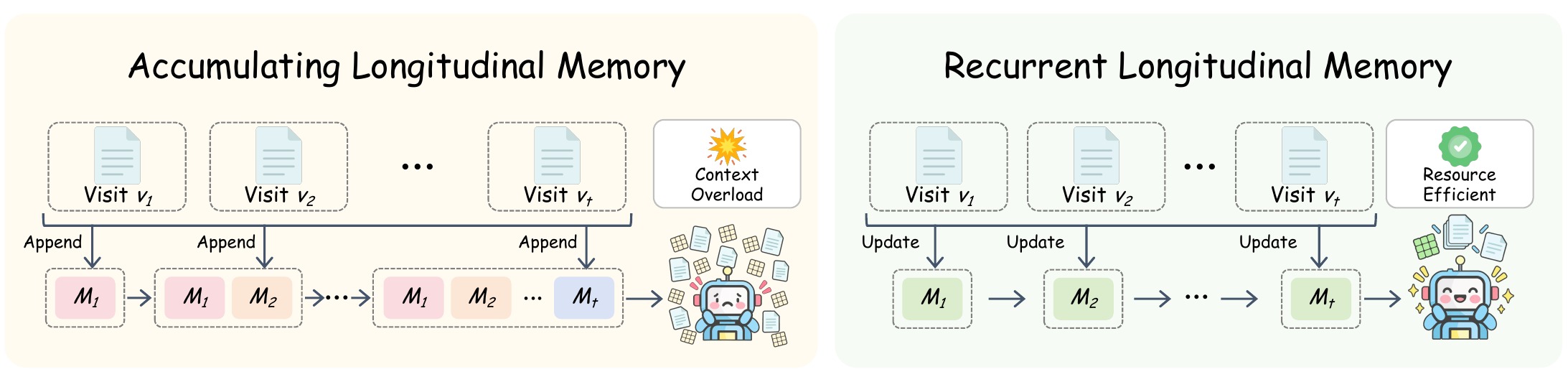}
    \caption{Comparison between accumulating and recurrent longitudinal memory. Left: Each new visit adds a separate memory block, potentially leading to context overload as visits accumulate. Right: ReLMem integrates each incoming visit with the previous memory to maintain a fixed-capacity state, enabling resource-efficient modeling.}
    \label{fig:memory-paradigm}
\end{figure}

Context compression offers a practical way to reduce these costs by replacing the full history with a compact representation. Existing methods typically compress a given context by pruning less informative KV entries or encoding its content into a smaller set of learned representations \citep{li2024snapkv,kim2025kvzip,ge2024icae}. But for longitudinal EHRs, compression must accommodate a history that continually expands with new visits. Recompressing the full history at each visit requires revisiting earlier records, whereas compressing visits separately and appending their representations still causes the retained context to grow \citep{kim2024ccm}. This motivates a recurrent formulation in which each incoming visit is integrated with the previous memory to produce an updated state of fixed capacity \citep{bulatov2022rmt}. As illustrated in Figure~\ref{fig:memory-paradigm}, this formulation maintains a continually updated patient memory without rereading earlier records or accumulating separate representations for successive visits.

However, maintaining such a memory requires more than just effectively compressing individual visits. Each update must incorporate new clinical information while preserving historical evidence that may be needed for downstream tasks \citep{rae2020compressive}. Since information from earlier visits is accessible only through the previous memory, information loss at one step can persist and accumulate across subsequent updates \citep{kim2024ccm}. Although supervision with ground-truth answers encourages accurate predictions from the final memory, it does not explicitly constrain how well intermediate memory states preserve historical evidence. This highlights a central challenge in efficient longitudinal EHR modeling: \textit{how to preserve task-relevant information throughout recurrent updates while ensuring that the resulting memory supports downstream clinical prediction.}

To address this challenge, we introduce \textbf{Re}current \textbf{L}ongitudinal \textbf{Mem}ory (\textbf{ReLMem}), a framework that learns fixed-capacity patient memory for longitudinal EHR modeling with a frozen, task-adapted LLM. At each visit, this LLM uses lightweight compression adapters to integrate the complete incoming record with the previous memory and replace it with an updated state. We develop a multi-granularity optimization strategy to preserve historical evidence throughout recurrent updates while supporting downstream prediction. Specifically, we align attention outputs from intermediate memory states with those from the complete available history under identical queries, and supervise predictions from the final memory using cross-entropy with ground truth answers. We further adopt a curriculum learning strategy that progressively introduces cases with more visits to help ReLMem maintain informative patient memory over longer histories. Across two representative longitudinal EHR tasks (i.e., medication and diagnosis prediction), ReLMem approaches the F1 scores of the full-history baseline while using a substantially smaller history budget.

Our contributions are summarized as follows:
\begin{itemize}
    \item We introduce ReLMem, a framework that learns fixed-capacity recurrent patient memory for resource efficient longitudinal EHR modeling.
    \item We develop multi-granularity optimization that aligns intermediate memory readouts with those from full history and jointly supervises clinical predictions from the final memory.
    \item We evaluate ReLMem on representative longitudinal EHR tasks, demonstrating a favorable trade-off between predictive performance and retained-history storage.
\end{itemize}

\section{Related work}
\paragraph{Longitudinal EHR modeling.}
Longitudinal EHR modeling integrates information across successive visits to support clinical prediction. Early neural models used patient histories for diagnosis, medication recommendation, and risk prediction~\citep{choi2016doctorai,choi2016retain,shang2019gamenet}. Pretrained transformers subsequently improved disease prediction and enabled few-shot adaptation~\citep{li2020behrt,rasmy2021medbert,wornow2023ehrshot}, with TransformEHR further demonstrating the benefit of more complete visit histories~\citep{yang2023transformehr}. More recently, generative models have been used to forecast clinical events and disease trajectories~\citep{kraljevic2024foresight,shmatko2025delphi}, while Apollo predicts disease progression and treatment response from multimodal records~\citep{zhang2026apollo}. EHR-R1 and EHRAgent further extend EHR analysis to LLM-based clinical reasoning and multi-step database querying, respectively~\citep{liao2025ehrr1,shi2024ehragent}. ReLMem complements these advances by learning recurrent patient memory that preserves task-relevant history for downstream prediction with a frozen LLM.

\paragraph{Context compression.}
Context compression condenses information for efficient processing and retention by LLMs. LLMLingua and LLMLingua-2 shorten prompts by removing less informative content~\citep{jiang2023llmlingua,pan2024llmlingua2}, whereas Gisting and the In-context Autoencoder encode text into compact learned representations~\citep{mu2023gisting,ge2024icae}. At the KV-cache level, H$_2$O, SnapKV, and KVzip reduce storage requirements by selectively retaining cached states~\citep{zhang2023h2o,li2024snapkv,kim2025kvzip}. More recently, Cartridges and Attention Matching directly optimize compact KV representations for a given input~\citep{eyuboglu2025cartridges,zweiger2026attentionmatching}. However, compressed representations can still accumulate across successive visits, potentially leading to context overload. Recurrent memory offers a practical alternative by integrating each incoming visit with the previous state to maintain resource efficient modeling.

\paragraph{Recurrent memory.}
Recurrent memory carries information across successive segments to support efficient longitudinal modeling. Transformer-XL extends context through segment-level recurrence~\citep{dai2019transformerxl}, while Compressive Transformer retains compressed historical states with attention reconstruction supervision~\citep{rae2020compressive}. Subsequent approaches learn compact representations to transfer or accumulate information across segments~\citep{bulatov2022rmt,chevalier2023autocompressor}. More recently, Compressed Context Memory (CCM) supports online interactions through recurrent compression~\citep{kim2024ccm}. For longitudinal EHR modeling, however, the key challenge of recurrent memory is not merely to compress individual visits but to integrate new information while preserving historical evidence for downstream prediction. To this end, ReLMem learns a replacement memory at each visit, aligning its attention outputs with those of an independently encoded complete historical prefix and jointly supervising predictions from the final memory.

\section{Problem Formulation}
\label{sec:problem}
Let $H_t=(v_1,\ldots,v_t)$ denote a patient's longitudinal history of completed visits in chronological order, where $v_t$ is the full text of the $t$-th visit. Given the history $H_T$ preceding a target visit and a task query $q$, the goal is to predict an answer sequence $y=(y_1,\ldots,y_N)$. Specifically, for medication prediction, $q$ includes the diagnoses and procedures of the target visit, and $y$ represents its medication set. For next-visit diagnosis prediction, $q$ contains only the task instruction, and $y$ represents the diagnoses at the next visit.

We consider longitudinal EHR modeling with a patient memory $M$ whose capacity remains fixed as visits accumulate. As each visit arrives, the LLM updates its memory from the previous state $M_{t-1}$ and the incoming visit $v_t$:
\begin{equation}
    M_0=\varnothing, \qquad
    M_t=\mathcal{C}_{\theta,\phi}(M_{t-1},v_t),
    \quad t=1,\ldots,T,
    \label{eq:recurrence}
\end{equation}
Here, $\mathcal{C}_{\theta,\phi}$ denotes the recurrent update performed by the LLM. The backbone parameters $\theta$ remain frozen, while $\phi$ denotes the additional parameters trained for memory compression. At inference, information from processed visits is accessible only through $M_t$. After the final update, the LLM predicts the answer according to $p_{\theta}(y\mid M_T,q)$. The objective of ReLMem is to learn recurrent memory updates that preserve task-relevant evidence for accurate clinical prediction.

\section{ReLMem}
\label{sec:method}

\begin{figure}[t]
    \centering
    \includegraphics[width=\textwidth]{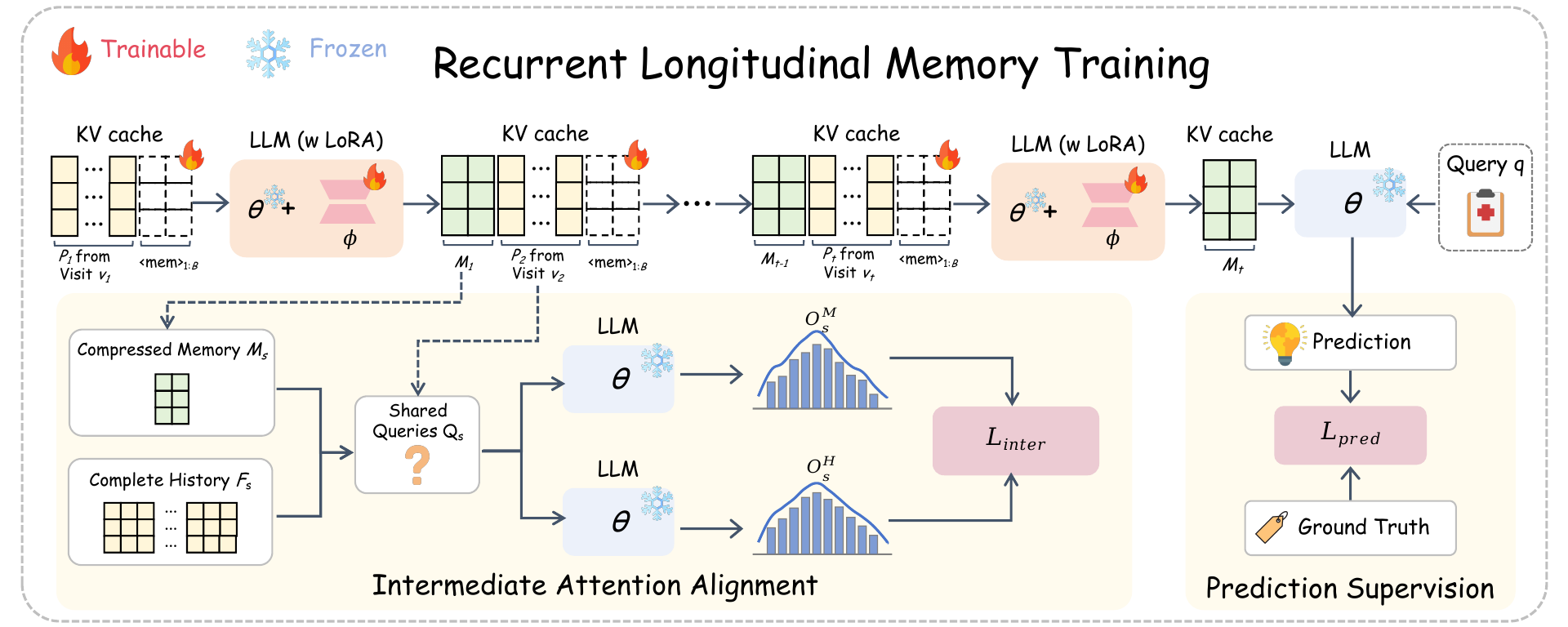}
    \caption{The training pipeline of recurrent longitudinal memory. A frozen LLM equipped with lightweight compression adapters encodes each incoming visit with the previous memory to produce an updated state of fixed capacity. $\mathcal{L}_{\mathrm{inter}}$ aligns attention readouts from current memory and the full history under identical queries, while $\mathcal{L}_{\mathrm{pred}}$ optimizes predictions from the final memory.}
    \label{fig:training-overview}
\end{figure}

ReLMem uses a task-adapted LLM backbone for recurrent memory updates and clinical prediction. We first adapt the LLM to the specific task using complete patient histories, then freeze the adapted backbone and train lightweight compression adapters to integrate each incoming visit with the previous memory. Recurrent memory learning uses multi-granularity optimization, combining intermediate attention alignment with final prediction supervision. We further adopt a curriculum learning strategy that starts with short histories and progressively introduces training examples with more visits. Figure~\ref{fig:training-overview} illustrates recurrent memory learning after task-specific adaptation.

\subsection{Task-Specific Adaptation}
\label{sec:task_adaptation}

To provide a task-specific backbone for recurrent memory learning, we first adapt the LLM to clinical prediction through supervised fine-tuning (SFT) on complete patient histories. Using low-rank adaptation (LoRA)~\citep{hu2021lora}, we keep the pretrained model $\theta_0$ fixed and optimize the task-adapter parameters $\psi$ by minimizing cross-entropy over the answer tokens:
\begin{equation}
    \mathcal{L}_{\mathrm{SFT}}(\psi)
    = \mathbb{E}_{(H_T,q,y)\sim\mathcal{D}}\!\left[
        -\frac{1}{N}\sum_{j=1}^{N}
        \log p_{\theta_0,\psi}(y_j\mid H_T,q,y_{<j})
    \right].
    \label{eq:sft}
\end{equation}
Here, $\mathcal{D}$ denotes the training set and $y_{<j}$ denotes the preceding ground-truth answer tokens. After adaptation, we merge the learned LoRA updates into the pretrained weights $\theta_0$ to obtain the backbone parameters $\theta$, which remain frozen throughout recurrent longitudinal memory training.

\subsection{Recurrent Memory Update}
\label{sec:memory_synthesis}

\begin{wrapfigure}{r}{.5\textwidth}
    \centering
    \includegraphics[width=0.5\textwidth]{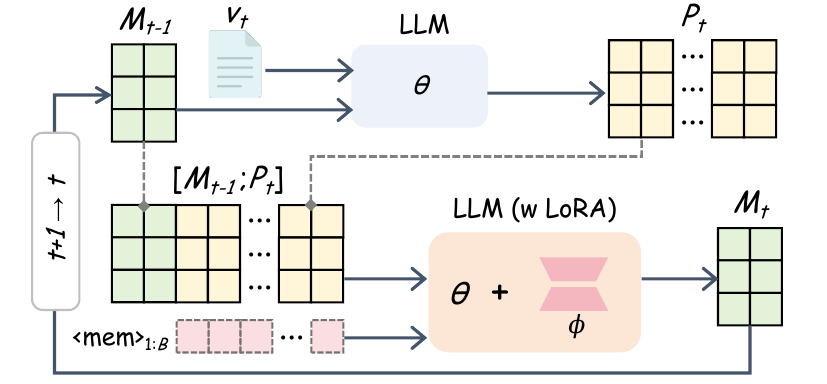}
    \caption{Recurrent memory update in ReLMem.}
    \label{fig:recurrent-memory-update}
\end{wrapfigure}

As shown in Figure~\ref{fig:recurrent-memory-update}, with the task-adapted backbone fixed, ReLMem updates patient memory in two steps: encoding each incoming visit in the context of the retained history, then integrating both into a new fixed-capacity state. To make the retained history directly accessible through the backbone's attention mechanism, we represent memory as layer-wise KV pairs:
\begin{equation}
    M_t=\bigl\{(\mathbf{K}_t^{\ell},\mathbf{V}_t^{\ell})\bigr\}_{\ell=1}^{L},
    \qquad
    \mathbf{K}_t^{\ell},\mathbf{V}_t^{\ell}
    \in\mathbb{R}^{H_{\mathrm{kv}}\times B\times d_h},
    \label{eq:memory_state}
\end{equation}
where $B$ is the fixed number of memory slots per KV head in each layer, and $L$, $H_{\mathrm{kv}}$, and $d_h$ denote the number of backbone layers, the number of KV heads, and the head dimension, respectively.

\paragraph{Memory-conditioned encoding.}
Rather than encoding each visit in isolation, the frozen backbone first profiles the full text of the incoming visit $v_t$ with $M_{t-1}$ as historical context:
\begin{equation}
    P_t=\operatorname{KV}_{\theta}(v_t\mid M_{t-1}),
    \label{eq:visit_encoding}
\end{equation}
where $\operatorname{KV}_{\theta}$ denotes KV computation through the frozen backbone. The resulting $P_t$ denotes the layer-wise KV pairs for all tokens in the current visit $v_t$.

\paragraph{Fixed-capacity update.}
We next incorporate the encoded visit into patient memory while keeping its capacity fixed. Specifically, we append a sequence of $B$ memory tokens, denoted by $\langle\mathrm{mem}\rangle_{1:B}$, to the end of the visit. Following CCM~\citep{kim2024ccm}, we use token-conditional low-rank adapters that are enabled only at memory-token positions, while visit, task-query, and answer tokens use the frozen backbone projections. The memory-token embeddings and adapter weights form the trainable parameters $\phi$, which are shared across all recurrent updates. The task-adapted backbone processes these memory tokens with $[M_{t-1};P_t]$ as a prefix KV cache, allowing them to attend to both the previous memory and the encoded visit:
\begin{equation}
    M_t=\operatorname{KV}_{\theta,\phi}
    \bigl(\langle\mathrm{mem}\rangle_{1:B}
    \mid[M_{t-1};P_t]\bigr),
    \label{eq:memory_synthesis}
\end{equation}
where $[\,;\,]$ denotes concatenation of KV states along the sequence dimension, and $\operatorname{KV}_{\theta,\phi}$ returns only the layer-wise KV pairs produced at the memory-token positions. We retain these KV pairs as $M_t$ in place of the previous memory and the temporary visit states. The updated memory provides historical context for the next visit, completing the recurrent update in Eq.~\ref{eq:recurrence}.

\subsection{Multi-Granularity Optimization}
\label{sec:optimization}
Such recurrent updates may discard information during compression~\citep{kim2024ccm}, and these losses can accumulate across visits and affect the final prediction. We therefore develop a multi-granularity optimization strategy that combines intermediate attention alignment to preserve historical evidence during updates with final prediction supervision to support accurate clinical prediction.

\paragraph{Intermediate attention alignment.}
To supervise the historical information retained in $M_s$ at any visit step $s\in\{1,\ldots,T\}$, we independently encode the complete history $H_s$ with the same frozen backbone to obtain an uncompressed reference $F_s=\operatorname{KV}_{\theta}(H_s)$. However, $M_s$ and $F_s$ contain different numbers of KV entries, preventing direct entry-wise comparison. We therefore align their attention outputs under identical queries, as these readouts reflect what the backbone retrieves from each representation.

Specifically, the frozen backbone processes the next visit $v_{s+1}$, or the task query $q$ when $s=T$, using $M_s$ as historical context. We then use the resulting attention query vectors $\mathbf{Q}_s$ to read from both representations:
\begin{equation}
    \begin{aligned}
        \mathbf{O}_s^M = \operatorname{Attn}(\mathbf{Q}_s,M_s), \qquad
        \mathbf{O}_s^H = \operatorname{Attn}(\mathbf{Q}_s,F_s),
    \end{aligned}
    \label{eq:attention_outputs}
\end{equation}
where $\operatorname{Attn}$ reads from the historical KV pairs in the supplied representation. For a single layer and query head, this operation is computed as:
\begin{equation}
    \operatorname{Attn}(\mathbf{Q},X)
    =
    \operatorname{softmax}\!\left(
        \frac{\mathbf{Q}\mathbf{K}_X^{\top}}{\sqrt{d_h}}
    \right)\mathbf{V}_X,
    \qquad X\in\{M_s,F_s\},
    \label{eq:attention_readout}
\end{equation}
where $\mathbf{K}_X$ and $\mathbf{V}_X$ are the corresponding historical keys and values stored in $X$. The shared queries yield outputs of the same shape, allowing direct comparison despite the different lengths of the two representations. Finally, we minimize their normalized squared difference, averaged across the layers and query heads used for alignment:
\begin{equation}
    \mathcal{L}_{\mathrm{inter}}(\phi;s)
    =
    \frac{1}{|\mathcal{S}|H_q}
    \sum_{\ell\in\mathcal{S}}
    \sum_{h=1}^{H_q}
    \frac{
        \left\|
            \mathbf{O}_{s,\ell,h}^{M}
            -
            \mathbf{O}_{s,\ell,h}^{H}
        \right\|_F^2
    }{
        \left\|
            \mathbf{O}_{s,\ell,h}^{H}
        \right\|_F^2
        + \epsilon n_s d_h
    },
    \label{eq:alignment_loss}
\end{equation}
where $\mathcal{S}$ denotes the layers used for alignment, $H_q$ is the number of query heads, $n_s$ is the number of query vectors per head, and $\epsilon>0$ stabilizes the normalization. By matching what the backbone retrieves from the full history, this objective encourages the updated memory to preserve historical context for subsequent visits and clinical prediction.

\paragraph{Final prediction supervision.}
While intermediate alignment supervises access to historical information, final prediction supervision directly targets the clinical answer. We condition the backbone on the final memory $M_T$ and task query $q$, and minimize cross-entropy over the ground-truth answer:
\begin{equation}
    \mathcal{L}_{\mathrm{pred}}(\phi)
    =
    -\frac{1}{N}
    \sum_{j=1}^{N}
    \log p_{\theta}
    \left(y_j\mid M_T,q,y_{<j}\right).
    \label{eq:prediction_loss}
\end{equation}
With $\theta$ fixed, this objective only optimizes $\phi$, guiding recurrent updates to retain information useful for clinical prediction. At last, we combine intermediate attention alignment and final prediction supervision in a joint objective:
\begin{equation}
    \min_{\phi}\;
    \mathbb{E}_{(H_T,q,y)\sim\mathcal{D}}
    \left[
        \mathcal{L}_{\mathrm{pred}}(\phi)
        +
        \frac{\lambda}{T}
        \sum_{s=1}^{T}
        \mathcal{L}_{\mathrm{inter}}(\phi;s)
    \right],
    \label{eq:joint_objective}
\end{equation}
where $\lambda$ controls the contribution of intermediate alignment relative to final prediction supervision. The full-history reference is used only during training.

\subsection{Curriculum Learning}
\label{sec:curriculum}

As the number of visits increases, information loss may accumulate over more recurrent updates, making memory learning more challenging. We therefore adopt a curriculum learning strategy~\citep{bengio2009curriculum} that begins with short histories and progressively introduces examples with more visits. At each training stage $e$, we draw examples from
\begin{equation}
    \mathcal{D}_e
    = \bigl\{(H_T,q,y)\in\mathcal{D}:T\leq\tau_e\bigr\},
    \label{eq:curriculum}
\end{equation}
where $\tau_e$ is a nondecreasing visit-count threshold. The eligible set gradually expands to cover the full training set, while short histories continue to be sampled alongside longer ones to maintain supervision across recurrence depths. Throughout this progression, the memory capacity, recurrent update rule, and joint objective in Eq.~\ref{eq:joint_objective} remain unchanged.

\section{Experiments}
\subsection{Datasets and evaluation protocol}
We evaluate ReLMem on medication and next-visit diagnosis prediction using MIMIC-IV~\citep{johnson2023mimiciv}. Following prior work~\citep{shang2019gamenet,wang2021sarmr}, this retrospective medication prediction uses completed visits and the target admission's diagnoses and procedures to predict medication classes prescribed within its first 24 hours of hospitalization. Diagnosis prediction instead infers the next visit's diagnosis categories from completed visits alone. For each task, the test set is derived from EHRs of 300 patients. We also construct corresponding training and validation sets for recurrent memory learning, with no patient overlap across the three splits. Appendices~\ref{app:data} and~\ref{app:diagnosis_generalization} provide further details. For downstream evaluation, we report macro-F1 averaged across cases and micro-F1 calculated over the test set as a whole. We also report P@$k$ and R@$k$ for $k \in \{5,10\}$, computed over the first $k$ unique labels in generation order. To assess resource efficiency, we measure retained-history storage, peak GPU memory, update latency, and prediction latency. Detailed metric definitions are provided in Appendix~\ref{app:evaluation}.

\subsection{Implementation details}
We use Qwen3-4B~\citep{yang2025qwen3} unless otherwise specified. For each task, we first adapt the backbone with LoRA~\citep{hu2021lora} for one epoch at a learning rate of $10^{-4}$. We then freeze the adapted backbone and train only the memory-token embeddings and compression adapters, with a default memory capacity of $B=1{,}024$ and alignment weight $\lambda=0.1$. To limit training overhead, we compute intermediate attention alignment at one uniformly sampled update boundary per example, using four layers distributed across model depth. Memory learning uses AdamW with a peak learning rate of $3\times10^{-4}$ for up to five epochs. The curriculum learning progressively introduces training examples with more visits, while early stopping is guided by answer cross-entropy on the validation set. All evaluated inputs fit within the configured context window, with space reserved for answer generation, except in the exploration analysis in appendix~\ref{app:extended_histories}. More details about training and inference of ReLMem can be found in appendix~\ref{app:training_details}.

\subsection{Baselines}
We compare ReLMem with Full History, an uncompressed reference, and six history compression methods. LLM-Rsum~\citep{wang2025rsum} recursively updates a textual summary as visits arrive. SnapKV~\citep{li2024snapkv}, KVzip~\citep{kim2025kvzip}, and Attention Matching~\citep{zweiger2026attentionmatching} are adapted to compress the previous memory together with each incoming visit. RMT~\citep{bulatov2022rmt} and CCM-merge~\citep{kim2024ccm} learn recurrent memory under their respective supervision schemes. All methods share the same task-adapted backbone. Appendix~\ref{app:baseline_details} provides more details on implementation of each baseline.

\subsection{Main results}
As shown in Table~\ref{tab:medication_results}, ReLMem approaches Full History using only 2.9\% of its average history budget and outperforms all evaluated memory compression methods. It achieves the best P@5 and R@5 scores, showing strong precision and reference-set coverage among the first five generated labels. Across baselines, KVzip substantially outperforms LLM-Rsum, showing the predictive value of retained KV states in longitudinal EHRs modeling. Similarly, RMT trails CCM-merge and ReLMem indicating a representational bottleneck in carrying history through final-layer embeddings rather than layer-wise KV states. Additionally, the weaker results of the SnapKV and Attention Matching adaptations highlight the difficulty of transferring direct context compression to recurrent updates. ReLMem's advantage over CCM-merge is also consistent with the benefit of intermediate supervision observed in the objective ablations (Table~\ref{tab:aba_loss}), which further highlights the value of multi-granularity optimization for recurrent memory learning.

\begin{table}[t]
    \centering
    \caption{Medication prediction results (\%) with ReLMem and baselines. The best and sub-optimal scores among history compression methods are shown in \textbf{bold} and \underline{underlined}, respectively.}
    \label{tab:medication_results}
    \resizebox{\textwidth}{!}{%
    \begin{tabular}{@{}lcccc r@{\,}l r@{\,}l r@{}}
        \toprule
        \textbf{Method}
        & \textbf{P@5}
        & \textbf{P@10}
        & \textbf{R@5}
        & \textbf{R@10}
        & \multicolumn{2}{c}{\textbf{Macro-F1 (95\% CI)}}
        & \multicolumn{2}{c}{\textbf{Micro-F1 (95\% CI)}}
        & \multicolumn{1}{c@{}}{\makecell[c]{\textbf{History}\\\textbf{Budget}}} \\
        \midrule

        \multicolumn{10}{@{}l}{\textit{Uncompressed reference}} \\
        Full History
        & 68.80 & 63.23 & 32.77 & 58.24
        & 62.22 & (60.46, 63.92)
        & 63.32 & (61.69, 64.88)
        & 35,279 \\
        \midrule

        \multicolumn{10}{@{}l}{\textit{History compression methods}} \\

        LLM-Rsum
        & 31.73 & 20.93 & 14.76 & 18.99
        & 18.97 & (17.53, 20.46)
        & 19.21 & (17.80, 20.69)
        & 1,026 \\

        SnapKV
        & 21.93 & 13.40 & 9.59 & 11.68
        & 11.91 & (10.62, 13.20)
        & 11.95 & (10.71, 13.25)
        & 1,024 \\

        KVzip
        & 63.73 & 54.03 & 30.06 & 49.44
        & 52.53 & (50.41, 54.59)
        & 54.52 & (52.73, 56.28)
        & 1,024 \\

        Attention Matching
        & 9.40 & 5.97 & 4.32 & 5.35
        & 6.44 & (5.13, 7.83)
        & 8.32 & (6.70, 9.99)
        & 1,024 \\

        RMT
        & 64.13 & 55.07 & 30.68 & 50.90
        & 47.28 & (45.65, 48.89)
        & 48.42 & (46.97, 49.88)
        & 1,024 \\

        CCM-merge
        & \underline{67.93}
        & \underline{61.60}
        & \underline{32.42}
        & \underline{56.61}
        & \underline{57.27} & (55.53, 58.95)
        & \underline{58.49} & (56.93, 59.99)
        & 1,024 \\

        \textbf{ReLMem (ours)}
        & \textbf{70.53}
        & \textbf{63.33}
        & \textbf{33.50}
        & \textbf{58.13}
        & \textbf{61.93} & (60.21, 63.64)
        & \textbf{63.24} & (61.62, 64.85)
        & 1,024 \\
        \bottomrule
    \end{tabular}%
    }
\end{table}

\subsection{Analysis}
\label{sec:analysis}

\subsubsection{Resource Costs of Recurrent Compression}
\label{sec:recurrent_cost}
In Table~\ref{tab:memory_overhead}, we compare ReLMem's inference costs with Full History: it nearly halves peak GPU memory and reduces average update latency by 26.3\%. Although Full History reuses its KV cache to avoid re-encoding earlier visits, processing each incoming visit still requires attention over all historical KV states, increasing inference costs as history grows. By operating on a compact historical state, ReLMem maintains and updates patient memory more efficiently in longitudinal EHRs, despite encoding each incoming visit in full and performing an additional compression step.

\begin{table}[t]
    \centering
    \begin{minipage}[c]{0.495\textwidth}
        \caption{Resource costs of longitudinal medication prediction.}
        \label{tab:memory_overhead}
        \resizebox{\textwidth}{!}{%
        \begin{tabular}{lrr}
            \toprule
            Resource & Full History & ReLMem \\
            \midrule
            Historical KV state (MiB) & 4,961.08 & 144.00 \\
            Update latency (s/visit) & 0.4676 & 0.3446 \\
            Prediction latency (s/query) & 8.0837 & 6.5293 \\
            Peak GPU memory (GiB) & 17.4354 & 9.2379 \\
            \bottomrule
        \end{tabular}
        }
    \end{minipage}
    \hfill
    \begin{minipage}[c]{0.49\textwidth}
        \caption{Effects of training objectives and curriculum learning.}
        \label{tab:aba_loss}
        \resizebox{\textwidth}{!}{%
        \begin{tabular}{ccc|cc}
            \toprule
            $\mathcal{L}_{\mathrm{inter}}$ & $\mathcal{L}_{\mathrm{pred}}$ & Curriculum & Macro-F1 & Micro-F1 \\
            \midrule
            $\checkmark$ & & $\checkmark$ & 0.37 & 0.55 \\
            & $\checkmark$ & $\checkmark$ & 47.67 & 48.65 \\
            $\checkmark$ & $\checkmark$ & & 61.00 & 62.24 \\
            $\checkmark$ & $\checkmark$ & $\checkmark$ & \textbf{61.93} & \textbf{63.24} \\
            \bottomrule
        \end{tabular}
        }
    \end{minipage}
\end{table}

\subsubsection{Memory-Performance Trade-offs}
\label{sec:memory_tradeoff}

\begin{figure}[t]
    \centering
    \includegraphics[width=\linewidth]{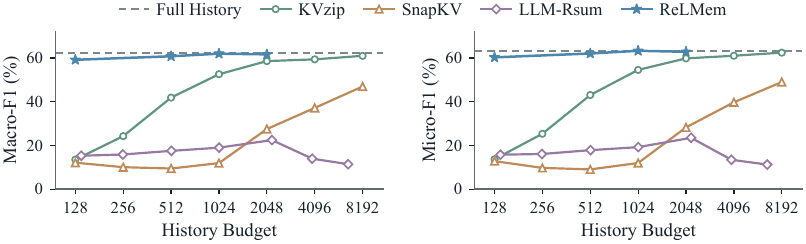}
    \caption{Medication prediction performance across different history budgets.}
    \label{fig:memory_tradeoff}
\end{figure}

We further analyze how the history budget affects performance (Figure~\ref{fig:memory_tradeoff}). Reducing ReLMem's budget eightfold, from 1,024 to 128 slots, lowers macro- and micro-F1 by only 2.81 and 3.00 percentage points, respectively. At 128 slots, it still outperforms KVzip at 2,048 positions on both metrics, using just one-sixteenth of the history budget. This suggests that learned recurrent states preserve predictive evidence more densely than selected KV entries. In contrast, LLM-Rsum peaks at 2,048 and declines thereafter, showing that a larger summary budget does not necessarily improve prediction. Overall, these results highlight the value of learning to preserve task-relevant information through recurrent updates rather than simply increasing memory capacity.

\subsubsection{Effect of Visits Count and History Length}
\label{sec:history_scaling}
\begin{figure}[t]
    \centering
    \includegraphics[width=\linewidth]{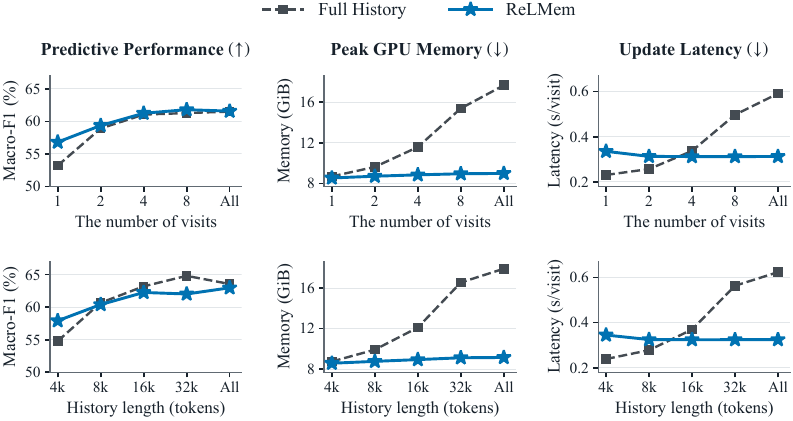}
    \caption{Predictive performance and computational efficiency of ReLMem and Full History across different numbers of historical visits (top) and history lengths (bottom).}
    \label{fig:history_scaling}
\end{figure}

Beyond the memory budget, we further compare ReLMem and Full History by varying the number of recent visits or the history length for each patient. We construct two cohorts of 110 and 107 cases from the medication test set for these respective analyses, keeping each prediction target fixed and varying only how much recent history is provided. As shown in Figure~\ref{fig:history_scaling}, both methods benefit from additional history, highlighting the value of earlier visits for downstream prediction. As more history is included, however, Full History's peak GPU memory more than doubles, whereas ReLMem's increases by only 5.3-6.6\%. With all available history, ReLMem remains within 0.59 percentage points of Full History in macro-F1 while reducing peak GPU memory by 48.8-49.1\% and final-visit update latency by 47.2-47.9\%. These results highlight ReLMem's growing efficiency advantage in longitudinal EHR modeling as patient histories expand.

\subsubsection{Contributions of Training Objectives and Curriculum Learning}
\label{sec:objective_analysis}
We next examine the contributions of ReLMem's training components by ablating the supervision objectives and curriculum learning (Table~\ref{tab:aba_loss}). Under the same curriculum, removing prediction supervision reduces both F1 scores to below 1\%, indicating that matching historical attention outputs alone does not ensure useful clinical predictions. Removing intermediate alignment instead lowers macro- and micro-F1 by 14.26 and 14.59 percentage points, respectively, highlighting the limitations of relying solely on final-answer supervision to guide information preservation across updates. Therefore, the two objectives are complementary: $\mathcal{L}_{\mathrm{inter}}$ encourages memory to retain key historical information, while $\mathcal{L}_{\mathrm{pred}}$ directs it toward the downstream clinical task. With both objectives retained, curriculum learning further improves performance, supporting the strategy of learning shorter update sequences before progressively handling longer histories.

\begin{table}[t]
    \centering
    \begin{minipage}[c]{0.59\textwidth}
        \caption{Average score of Macro-F1 and Micro-F1 on medication prediction across model scales and families.}
        \label{tab:model_sizes}
        \resizebox{\textwidth}{!}{
        \begin{tabular}{l|ccc}
            \toprule
            Method & Qwen3-4B & Qwen3-8B & Llama-3.1-8B \\
            \midrule
            Full History & 62.77 & 62.69 & 65.62 \\
            ReLMem       & 62.59 & 61.01 & 63.16 \\
            \bottomrule
        \end{tabular}
        }
    \end{minipage}
    \hfill
    \begin{minipage}[c]{0.4\textwidth}
        \caption{Next-visit diagnosis prediction with Qwen3-8B.}
        \label{tab:diagnosis}
        \resizebox{\textwidth}{!}{
        \begin{tabular}{lcc}
            \toprule
            Method & Macro-F1 & Micro-F1 \\
            \midrule
            Full History & 36.30 & 37.41 \\
            ReLMem       & 35.43 & 35.85 \\
            \bottomrule
        \end{tabular}
        }        
    \end{minipage}
\end{table}

\subsubsection{Evaluation across Model Scales and Families}
\label{sec:backbone_analysis}
Beyond Qwen3-4B, we further evaluate the scalability of ReLMem across model scales and families by extending it to Qwen3-8B~\citep{yang2025qwen3} and Llama-3.1-8B~\citep{grattafiori2024llama}, as shown in Table~\ref{tab:model_sizes}. Across all three backbones, the gap between ReLMem and Full History remains below 2.5 percentage points in average F1 score. This consistent performance demonstrates that ReLMem can be applied across different LLM backbones while preserving the predictive utility.

\subsubsection{Evaluation on Diagnosis Prediction}
\label{sec:diagnosis_analysis}
Additionally, we evaluate ReLMem on next-visit diagnosis prediction, where no information from the target visit is available and prediction relies entirely on longitudinal history. As shown in Table~\ref{tab:diagnosis}, ReLMem remains close to Full History, with a Macro-F1 gap of only 0.87 percentage points. This shows that the recurrent memory preserves useful historical context even when downstream prediction cannot rely on current-visit diagnoses or procedures.

\section{Conclusion}
We introduced ReLMem, a framework that learns fixed-capacity recurrent memory for longitudinal EHR modeling with a frozen LLM. With multi-granularity optimization, ReLMem approaches full history performance on two representative tasks and substantially reduces storage and latency as histories expand. Through extensive experiments, we demonstrate the importance of learning not just to compress patient history, but to preserve its predictive value across recurrent updates.

\section*{AI use statement}
LLMs were used for language refinement, formatting checks, and code debugging. All technical ideas, study design, implementation, experimental analysis, and scientific conclusions were developed and verified by the authors.

\section*{Ethics statement}
This study uses de-identified records from MIMIC-IV and MIMIC-IV-Note~\citep{johnson2023mimiciv,johnson2024mimiciv31,johnson2023mimicivnote}, whose access is governed by PhysioNet credentialing and data-use agreements. All experiments are retrospective and have no effect on patient care. The targets reflect recorded diagnoses and prescriptions, which may contain documentation and treatment biases. Clinical use would require prospective validation and clinician oversight.

\section*{Reproducibility statement}
Appendices~\ref{app:data} and~\ref{app:diagnosis_generalization} describe the procedure of dataset construction for both tasks. Appendix~\ref{app:method_details} presents the learning and inference algorithm and position-encoding strategy. Appendices~\ref{app:training_details} and~\ref{app:baseline_details} detail training, checkpoint selection, decoding, and baseline implementations, while Appendix~\ref{app:evaluation} defines the predictive metrics, resource measurements, and statistical analysis. Access to the underlying clinical records remains subject to the original dataset agreements. Code will be publicly released upon acceptance.

\bibliographystyle{iclr2027_conference}
\bibliography{reference}

\clearpage

\appendix

\begin{center}
    \large\bfseries Contents of the Appendix
\end{center}
\medskip
\begingroup
\small
\hypersetup{pdfborder={0 0 0}}
\setlength{\parindent}{0pt}
\setlength{\parskip}{2pt}
\textcolor{blue!55!black}{\hyperref[app:data]{\makebox[1.8em][l]{\ref*{app:data}}\nameref*{app:data}}}\dotfill\pageref{app:data}\par
\hspace*{1.4em}\textcolor{blue!55!black}{\hyperref[app:contents_1_1]{\makebox[2.5em][l]{\ref*{app:contents_1_1}}\nameref*{app:contents_1_1}}}\dotfill\pageref{app:contents_1_1}\par
\hspace*{1.4em}\textcolor{blue!55!black}{\hyperref[app:contents_1_4]{\makebox[2.5em][l]{\ref*{app:contents_1_4}}\nameref*{app:contents_1_4}}}\dotfill\pageref{app:contents_1_4}\par
\hspace*{1.4em}\textcolor{blue!55!black}{\hyperref[app:contents_1_2]{\makebox[2.5em][l]{\ref*{app:contents_1_2}}\nameref*{app:contents_1_2}}}\dotfill\pageref{app:contents_1_2}\par
\hspace*{1.4em}\textcolor{blue!55!black}{\hyperref[app:contents_1_3]{\makebox[2.5em][l]{\ref*{app:contents_1_3}}\nameref*{app:contents_1_3}}}\dotfill\pageref{app:contents_1_3}\par
\textcolor{blue!55!black}{\hyperref[app:diagnosis_generalization]{\makebox[1.8em][l]{\ref*{app:diagnosis_generalization}}\nameref*{app:diagnosis_generalization}}}\dotfill\pageref{app:diagnosis_generalization}\par
\hspace*{1.4em}\textcolor{blue!55!black}{\hyperref[app:contents_8_1]{\makebox[2.5em][l]{\ref*{app:contents_8_1}}\nameref*{app:contents_8_1}}}\dotfill\pageref{app:contents_8_1}\par
\hspace*{1.4em}\textcolor{blue!55!black}{\hyperref[app:contents_8_4]{\makebox[2.5em][l]{\ref*{app:contents_8_4}}\nameref*{app:contents_8_4}}}\dotfill\pageref{app:contents_8_4}\par
\hspace*{1.4em}\textcolor{blue!55!black}{\hyperref[app:contents_8_2]{\makebox[2.5em][l]{\ref*{app:contents_8_2}}\nameref*{app:contents_8_2}}}\dotfill\pageref{app:contents_8_2}\par
\hspace*{1.4em}\textcolor{blue!55!black}{\hyperref[app:contents_8_3]{\makebox[2.5em][l]{\ref*{app:contents_8_3}}\nameref*{app:contents_8_3}}}\dotfill\pageref{app:contents_8_3}\par
\textcolor{blue!55!black}{\hyperref[app:method_details]{\makebox[1.8em][l]{\ref*{app:method_details}}\nameref*{app:method_details}}}\dotfill\pageref{app:method_details}\par
\hspace*{1.4em}\textcolor{blue!55!black}{\hyperref[app:relmem_algorithm]{\makebox[2.5em][l]{\ref*{app:relmem_algorithm}}\nameref*{app:relmem_algorithm}}}\dotfill\pageref{app:relmem_algorithm}\par
\hspace*{1.4em}\textcolor{blue!55!black}{\hyperref[app:position_encoding]{\makebox[2.5em][l]{\ref*{app:position_encoding}}\nameref*{app:position_encoding}}}\dotfill\pageref{app:position_encoding}\par
\textcolor{blue!55!black}{\hyperref[app:training_details]{\makebox[1.8em][l]{\ref*{app:training_details}}\nameref*{app:training_details}}}\dotfill\pageref{app:training_details}\par
\hspace*{1.4em}\textcolor{blue!55!black}{\hyperref[app:contents_3_1]{\makebox[2.5em][l]{\ref*{app:contents_3_1}}\nameref*{app:contents_3_1}}}\dotfill\pageref{app:contents_3_1}\par
\hspace*{1.4em}\textcolor{blue!55!black}{\hyperref[app:contents_3_2]{\makebox[2.5em][l]{\ref*{app:contents_3_2}}\nameref*{app:contents_3_2}}}\dotfill\pageref{app:contents_3_2}\par
\hspace*{1.4em}\textcolor{blue!55!black}{\hyperref[app:optimization_dynamics]{\makebox[2.5em][l]{\ref*{app:optimization_dynamics}}\nameref*{app:optimization_dynamics}}}\dotfill\pageref{app:optimization_dynamics}\par
\hspace*{1.4em}\textcolor{blue!55!black}{\hyperref[app:contents_3_3]{\makebox[2.5em][l]{\ref*{app:contents_3_3}}\nameref*{app:contents_3_3}}}\dotfill\pageref{app:contents_3_3}\par
\textcolor{blue!55!black}{\hyperref[app:baseline_details]{\makebox[1.8em][l]{\ref*{app:baseline_details}}\nameref*{app:baseline_details}}}\dotfill\pageref{app:baseline_details}\par
\hspace*{1.4em}\textcolor{blue!55!black}{\hyperref[app:baseline_full_history]{\makebox[2.5em][l]{\ref*{app:baseline_full_history}}\nameref*{app:baseline_full_history}}}\dotfill\pageref{app:baseline_full_history}\par
\hspace*{1.4em}\textcolor{blue!55!black}{\hyperref[app:baseline_rsum]{\makebox[2.5em][l]{\ref*{app:baseline_rsum}}\nameref*{app:baseline_rsum}}}\dotfill\pageref{app:baseline_rsum}\par
\hspace*{1.4em}\textcolor{blue!55!black}{\hyperref[app:baseline_snapkv]{\makebox[2.5em][l]{\ref*{app:baseline_snapkv}}\nameref*{app:baseline_snapkv}}}\dotfill\pageref{app:baseline_snapkv}\par
\hspace*{1.4em}\textcolor{blue!55!black}{\hyperref[app:baseline_kvzip]{\makebox[2.5em][l]{\ref*{app:baseline_kvzip}}\nameref*{app:baseline_kvzip}}}\dotfill\pageref{app:baseline_kvzip}\par
\hspace*{1.4em}\textcolor{blue!55!black}{\hyperref[app:baseline_am]{\makebox[2.5em][l]{\ref*{app:baseline_am}}\nameref*{app:baseline_am}}}\dotfill\pageref{app:baseline_am}\par
\hspace*{1.4em}\textcolor{blue!55!black}{\hyperref[app:baseline_rmt]{\makebox[2.5em][l]{\ref*{app:baseline_rmt}}\nameref*{app:baseline_rmt}}}\dotfill\pageref{app:baseline_rmt}\par
\hspace*{1.4em}\textcolor{blue!55!black}{\hyperref[app:baseline_ccm]{\makebox[2.5em][l]{\ref*{app:baseline_ccm}}\nameref*{app:baseline_ccm}}}\dotfill\pageref{app:baseline_ccm}\par
\textcolor{blue!55!black}{\hyperref[app:evaluation]{\makebox[1.8em][l]{\ref*{app:evaluation}}\nameref*{app:evaluation}}}\dotfill\pageref{app:evaluation}\par
\hspace*{1.4em}\textcolor{blue!55!black}{\hyperref[app:contents_5_1]{\makebox[2.5em][l]{\ref*{app:contents_5_1}}\nameref*{app:contents_5_1}}}\dotfill\pageref{app:contents_5_1}\par
\hspace*{1.4em}\textcolor{blue!55!black}{\hyperref[app:contents_5_3]{\makebox[2.5em][l]{\ref*{app:contents_5_3}}\nameref*{app:contents_5_3}}}\dotfill\pageref{app:contents_5_3}\par
\hspace*{1.4em}\textcolor{blue!55!black}{\hyperref[app:statistical_analysis]{\makebox[2.5em][l]{\ref*{app:statistical_analysis}}\nameref*{app:statistical_analysis}}}\dotfill\pageref{app:statistical_analysis}\par
\textcolor{blue!55!black}{\hyperref[app:supplementary]{\makebox[1.8em][l]{\ref*{app:supplementary}}\nameref*{app:supplementary}}}\dotfill\pageref{app:supplementary}\par
\hspace*{1.4em}\textcolor{blue!55!black}{\hyperref[app:alignment_count]{\makebox[2.5em][l]{\ref*{app:alignment_count}}\nameref*{app:alignment_count}}}\dotfill\pageref{app:alignment_count}\par
\hspace*{1.4em}\textcolor{blue!55!black}{\hyperref[app:contents_6_5]{\makebox[2.5em][l]{\ref*{app:contents_6_5}}\nameref*{app:contents_6_5}}}\dotfill\pageref{app:contents_6_5}\par
\hspace*{1.4em}\textcolor{blue!55!black}{\hyperref[app:contents_6_6]{\makebox[2.5em][l]{\ref*{app:contents_6_6}}\nameref*{app:contents_6_6}}}\dotfill\pageref{app:contents_6_6}\par
\hspace*{1.4em}\textcolor{blue!55!black}{\hyperref[app:extended_histories]{\makebox[2.5em][l]{\ref*{app:extended_histories}}\nameref*{app:extended_histories}}}\dotfill\pageref{app:extended_histories}\par
\textcolor{blue!55!black}{\hyperref[app:qualitative]{\makebox[1.8em][l]{\ref*{app:qualitative}}\nameref*{app:qualitative}}}\dotfill\pageref{app:qualitative}\par
\hspace*{1.4em}\textcolor{blue!55!black}{\hyperref[app:cases_medication]{\makebox[2.5em][l]{\ref*{app:cases_medication}}\nameref*{app:cases_medication}}}\dotfill\pageref{app:cases_medication}\par
\hspace*{1.4em}\textcolor{blue!55!black}{\hyperref[app:cases_diagnosis]{\makebox[2.5em][l]{\ref*{app:cases_diagnosis}}\nameref*{app:cases_diagnosis}}}\dotfill\pageref{app:cases_diagnosis}\par
\textcolor{blue!55!black}{\hyperref[app:limitations]{\nameref*{app:limitations}}}\dotfill\pageref{app:limitations}\par
\endgroup
\clearpage

\section{Medication Prediction Dataset}
\label{app:data}

\subsection{Cohort Split and Target Construction}
\label{app:contents_1_1}
We construct medication prediction examples from MIMIC-IV~\citep{johnson2023mimiciv,johnson2024mimiciv31,johnson2023mimicivnote}. Each example combines completed hospital visits with the target visit's diagnoses and procedures to predict its medication classes. We include adult patients with prior hospital visits and a nonempty medication target, and split patients into disjoint training, validation, and test sets. Following GAMENet~\citep{shang2019gamenet} and SARMR~\citep{wang2021sarmr}, we define medication targets within the first 24 hours of hospitalization, which avoids combining treatment changes over hospital stays of different lengths. The 24-hour restriction only applies to the medication targets, while diagnoses and procedures are admission-level records used for retrospective prediction. Eligible prescriptions are marked as main medications and have valid start times at or after admission and before the earlier of the 24-hour boundary and discharge. Administration adjuncts such as flushes are excluded.

Following prior works, we map drug identifiers and names to ATC level-3 classes using a fixed ingredient-to-ATC crosswalk and collapse duplicate classes within each target set \citep{shang2019gamenet,yang2021safedrug,bodenreider2014rxclass}. To reduce label noise from incomplete normalization, we retain targets only when at least 90\% of both prescription rows and ingredient components can be mapped. Consistent with longitudinal medication recommendation, the history contains only completed visits available before the target admission, including their medication fields \citep{shang2019gamenet,wang2021sarmr}. All methods receive the same retained history and current-visit input.

\subsection{Clinical Record and Input Template}
\label{app:contents_1_4}
Table~\ref{tab:app_record_conventions} defines the completed-visit record shared by both tasks. It contains demographics, diagnoses, medications, clinical notes, procedures, and relative timing. The field names and types are fixed, while clinical values, list lengths, and the number of visits vary by patient. Table~\ref{tab:app_medication_prompt} specifies the prompt template used for medication prediction. The template consists of chronologically ordered visit records, a current-visit object containing only \texttt{diagnoses} and \texttt{procedures}, and a fixed task instruction defining the prediction target and required output format.

\begin{table}[!htbp]
    \centering
    \caption{Structure and fields of a completed-visit record shared by both tasks.}
    \label{tab:app_record_conventions}
    \small
    \begin{tabularx}{\linewidth}{@{}p{0.54\linewidth}@{\hspace{1em}}X@{}}
        \toprule
        \textbf{JSON record template} & \textbf{Field meanings} \\
        \midrule
        \begin{minipage}[t]{\linewidth}\ttfamily\raggedright
\{\par
\hspace*{0.96em}\char34{}demographics\char34{}: \{\par
\hspace*{1.92em}\char34{}age\char34{}: \textnormal{\textit{\textless{}age in years\textgreater{}}}, \char34{}sex\char34{}: \char34{}\textnormal{\textit{\textless{}sex\textgreater{}}}\char34{}\par
\hspace*{0.96em}\},\par
\hspace*{0.96em}\char34{}diagnoses\char34{}: [\char34{}\textnormal{\textit{\textless{}diagnosis\textgreater{}}}\char34{}, ...],\par
\hspace*{0.96em}\char34{}medications\char34{}: [\char34{}\textnormal{\textit{\textless{}medication\textgreater{}}}\char34{}, ...],\par
\hspace*{0.96em}\char34{}notes\char34{}: [\par
\hspace*{1.92em}\{\par
\hspace*{2.88em}\char34{}available\_day\char34{}: \textnormal{\textit{\textless{}availability day\textgreater{}}},\par
\hspace*{2.88em}\char34{}chart\_day\char34{}: \textnormal{\textit{\textless{}chart day\textgreater{}}},\par
\hspace*{2.88em}\char34{}note\_type\char34{}: \char34{}\textnormal{\textit{\textless{}note type\textgreater{}}}\char34{},\par
\hspace*{2.88em}\char34{}text\char34{}: \char34{}\textnormal{\textit{\textless{}note text\textgreater{}}}\char34{}\par
\hspace*{1.92em}\}, ...\par
\hspace*{0.96em}],\par
\hspace*{0.96em}\char34{}procedures\char34{}: [\char34{}\textnormal{\textit{\textless{}procedure\textgreater{}}}\char34{}, ...],\par
\hspace*{0.96em}\char34{}timeline\char34{}: \{\par
\hspace*{1.92em}\char34{}admit\_day\char34{}: \textnormal{\textit{\textless{}admission day\textgreater{}}},\par
\hspace*{1.92em}\char34{}available\_day\char34{}: \textnormal{\textit{\textless{}availability day\textgreater{}}},\par
\hspace*{1.92em}\char34{}discharge\_day\char34{}: \textnormal{\textit{\textless{}discharge day\textgreater{}}},\par
\hspace*{1.92em}\char34{}gap\_days\char34{}: \textnormal{\textit{\textless{}interval or null\textgreater{}}}\par
\hspace*{0.96em}\},\par
\hspace*{0.96em}\char34{}visit\_number\char34{}: \textnormal{\textit{\textless{}local visit index\textgreater{}}}\par
\}\par
        \end{minipage}
        &
        \begin{minipage}[t]{\linewidth}\raggedright
        \texttt{demographics}: age in years and recorded sex.\par\medskip
        \texttt{diagnoses}, \texttt{procedures}, and \texttt{medications}: lists of clinical names from this completed visit.\par\medskip
        \texttt{notes}: a variable-length list of notes. Each note contains its type, text, chart day, and availability day.\par\medskip
        \texttt{timeline}: admission, discharge, and availability days for the visit. \texttt{gap\_days} measures the interval since the preceding visible discharge.\par\medskip
        \texttt{visit\_number}: the position $1,\ldots,T$ in the visible history.\par\medskip
        All day values share the first visible admission as Day 0. The first visit has \texttt{gap\_days=null}. Clinical lists and \texttt{notes} can be empty (\texttt{[]}).
        \end{minipage} \\
        \bottomrule
    \end{tabularx}
\end{table}

\begin{table}[!htbp]
    \centering
    \caption{Input template for medication prediction.}
    \label{tab:app_medication_prompt}
    \small
    \begin{tabularx}{\linewidth}{@{}X@{}}
        \toprule
        \rowcolor{gray!10}\textbf{Completed history $H_T$} \\
        \textit{\textless{}Completed visit 1\textgreater{}}\par
        $\cdots$\par
        \textit{\textless{}Completed visit $T$\textgreater{}}\tabularnewline
        \midrule
        \rowcolor{gray!10}\textbf{Current visit} \\
        \raggedright
        Current visit diagnoses and procedures:\par
        {\ttfamily\{\char34{}diagnoses\char34{}:[\char34{}\textit{\textless{}diagnosis 1\textgreater{}}\char34{}, ...],\par
        \hspace*{0.6em}\char34{}procedures\char34{}:[\char34{}\textit{\textless{}procedure 1\textgreater{}}\char34{}, ...]\}}\tabularnewline
        \midrule
        \rowcolor{gray!10}\textbf{Task instruction} \\
        \raggedright
        Prediction task:\par
        Using the completed visits and the current visit diagnoses and procedures above, predict the complete set of ATC level-3 medication classes represented by qualifying prescriptions started during the first 24 hours of the current hospitalization. Historical medication fields describe their completed visits; they are not the current target.\par\medskip
        Record conventions:\par
        - Completed visits are ordered from earliest to latest.\par
        - admit\_day, discharge\_day, chart\_day, and available\_day are elapsed days from the admission of the first visible completed visit, which is Day 0; larger values are later.\par
        - visit\_number is local to this visible history. gap\_days is the interval from the previous visible discharge to the current admission and is null for the first visible visit.\par
        - timeline.available\_day is when all displayed information for a completed visit is treated as available. A note's chart\_day and available\_day use the same Day-0 origin.\par\medskip
        Output granularity: use ATC level-3 class names, not individual drugs, ingredients, brands, drug groups at another level, or ATC codes.\par\medskip
        Output format: return exactly one valid JSON object and nothing else:\par
        \{``predictions'':[``\textless{}medication 1\textgreater{}'',``\textless{}medication 2\textgreater{}'']\}\par\medskip
        Return the complete predicted set rather than a fixed top-K list. Do not include explanations, Markdown, or any additional key.\tabularnewline
        \bottomrule
    \end{tabularx}
\end{table}

\clearpage
\subsection{Dataset Statistics}
\label{app:contents_1_2}
\begin{figure}[!htb]
    \centering
    \includegraphics[width=\linewidth]{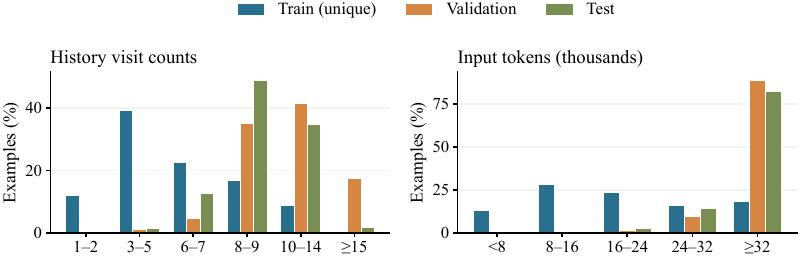}
    \caption{The distribution of historical visit counts and input tokens for each dataset.}
    \label{fig:medication_dataset}
\end{figure}

Table~\ref{tab:app_medication_statistics} summarizes cohort size and per-record statistics across the training, validation, and test splits. The same 6,804 training examples are used for both task adaptation and memory learning. Figure~\ref{fig:medication_dataset} shows the distributions of historical visit counts and input lengths. These dimensions impose complementary demands on recurrent memory: visit count determines recurrence depth, whereas input length reflects the volume of clinical information to be compressed. Compared with the training set, the test set contains more visits and longer contexts, providing a more demanding evaluation of whether fixed-capacity memory can preserve predictive evidence across recurrent updates.

\begin{table}[htbp]
    \centering
    \caption{Cohort, history, and target statistics for medication prediction datasets.}
    \label{tab:app_medication_statistics}
    \small
    \begin{tabular}{lccc}
        \toprule
        Statistic & Train & Validation & Test \\
        \midrule
        Patients & 1,452 & 75 & 300 \\
        Data items & 6,804 & 261 & 300 \\
        \midrule
        \multicolumn{4}{@{}l}{\textit{Per-record statistics: mean (min-max)}} \\
        History visits & 6.64 (2-21) & 11.28 (4-25) & 9.20 (3-19) \\
        Input tokens & 24,704 (692-40,449) & 36,760 (20,266-40,327) & 35,699 (20,537-40,335) \\
        Medication classes & 11.12 (1-28) & 12.69 (1-28) & 11.60 (1-28) \\
        \bottomrule
    \end{tabular}
\end{table}

\clearpage
\subsection{Label Distribution}
\label{app:contents_1_3}
\begin{figure}[htb]
    \centering
    \includegraphics[width=\linewidth]{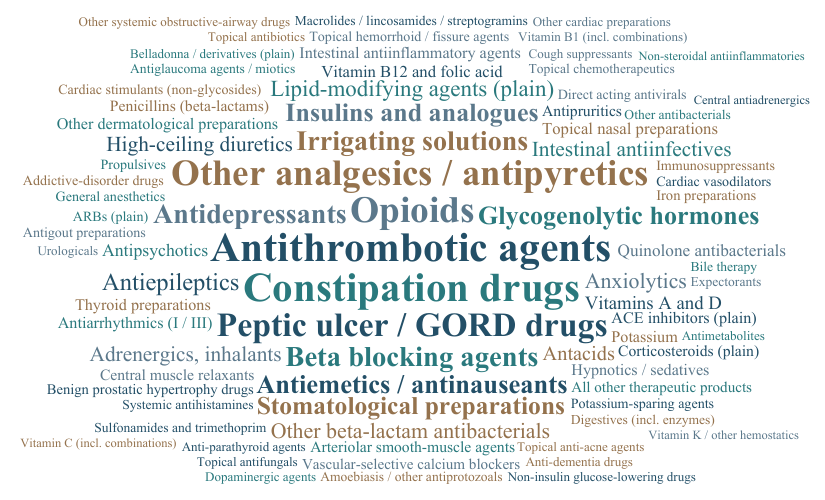}
    \caption{Word cloud of the most frequent medication classes in the test set.}
    \label{fig:medication_labels}
\end{figure}

The medication test set covers 107 classes across 300 target visits. Figure~\ref{fig:medication_labels} shows the 80 most frequent classes, each counted once per target visit. The distribution includes both common supportive treatments and less frequent therapeutic classes. Because target-set size varies across visits, the model must generate a variable-size medication set for each case.

\clearpage
\section{Diagnosis Prediction Dataset}
\label{app:diagnosis_generalization}

\subsection{Cohort Split and Target Construction}
\label{app:contents_8_1}
To explore the generalization capability for other tasks of ReLMem, we further construct a multi-label next-visit diagnosis prediction dataset from MIMIC-IV
\citep{johnson2023mimiciv,johnson2024mimiciv31,johnson2023mimicivnote}. Following the established longitudinal prediction setting \citep{choi2016doctorai}, each example uses a patient's completed hospital visits to predict the diagnosis categories documented at the subsequent admission, with all information from that target visit withheld. We split patients, rather than visits, into disjoint training, validation, and test cohorts to prevent patient overlap across splits. Patients may contribute multiple data items within the training and validation cohorts, whereas the test cohort contains one target per patient. During the process of construction, we normalize diagnosis codes to category-level labels using fixed ICD dictionaries. Following the category structures of ICD-9-CM and ICD-10-CM, category keys are formed from the first three characters of ICD-9-CM codes, except external-cause codes beginning with \texttt{E}, which use the first four characters, and from the first three characters of ICD-10-CM codes
\citep{nchs2011icd9cm,nchs2026icd10cm}. Each category is mapped to its standard title, and duplicate labels within the same target visit are removed.

\subsection{Clinical Record and Input Template}
\label{app:contents_8_4}

Diagnosis prediction uses the completed-visit records defined in Table~\ref{tab:app_record_conventions}. Table~\ref{tab:app_diagnosis_prompt} specifies the input template, which consists of chronologically ordered visit records followed by a fixed task instruction defining the prediction target and required output format. The template contains no current-visit object. The model returns a variable-length list of diagnosis-category titles under \texttt{predictions}, using the same JSON output format as medication prediction.

\begin{table}[htbp]
    \centering
    \caption{Input prompt for next-visit diagnosis prediction.}
    \label{tab:app_diagnosis_prompt}
    \small
    \begin{tabularx}{\linewidth}{@{}X@{}}
        \toprule
        \rowcolor{gray!10}\textbf{Completed history $H_T$} \\
        \textit{\textless{}Completed visit 1\textgreater{}}\par
        $\cdots$\par
        \textit{\textless{}Completed visit $T$\textgreater{}}\tabularnewline
        \midrule
        \rowcolor{gray!10}\textbf{Task instruction} \\
        \raggedright
        Prediction task:\par
        Using only the completed visits above, predict the complete set of diagnosis categories most likely to be documented in the patient's next hospital visit. No information from the target visit is provided.\par\medskip
        Record conventions:\par
        - Completed visits are ordered from earliest to latest.\par
        - admit\_day, discharge\_day, chart\_day, and available\_day are elapsed days from the admission of the first visible completed visit, which is Day 0; larger values are later.\par
        - visit\_number is local to this visible history. gap\_days is the interval from the previous visible discharge to the current admission and is null for the first visible visit.\par
        - timeline.available\_day is when all displayed information for a completed visit is treated as available. A note's chart\_day and available\_day use the same Day-0 origin.\par\medskip
        Output granularity: use standard ICD diagnosis-category titles. Each prediction should be broader than a specific clinical subtype and narrower than an organ-system grouping. Do not output ICD codes.\par\medskip
        Output format: return exactly one valid JSON object and nothing else:\par
        \{``predictions'':[``\textless{}diagnosis 1\textgreater{}'',``\textless{}diagnosis 2\textgreater{}'']\}\par\medskip
        Return the complete predicted set rather than a fixed top-K list. Do not include explanations, Markdown, or any additional key.\tabularnewline
        \bottomrule
    \end{tabularx}
\end{table}

\subsection{Dataset Statistics}
\label{app:contents_8_2}
\begin{figure}[!htb]
    \centering
    \includegraphics[width=\linewidth]{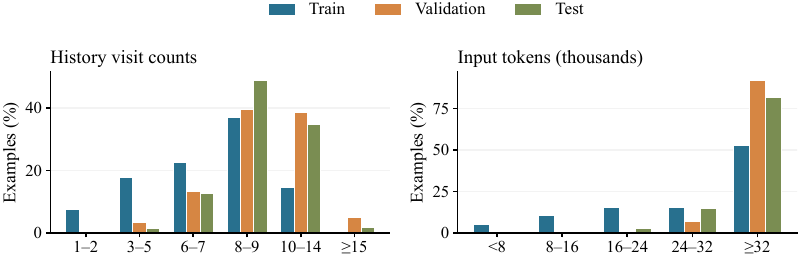}
    \caption{Diagnosis dataset characteristics: history visit counts and input lengths.}
    \label{fig:diagnosis_dataset}
\end{figure}

Table~\ref{tab:app_diagnosis_data} summarizes cohort size and per-record statistics across the training, validation, and test splits. The same 8,023 training examples are used for both task adaptation and memory learning. The test set contains one target visit per patient. Input lengths include the completed visits and task instruction, excluding the answer. Figure~\ref{fig:diagnosis_dataset} shows the distributions of history visit counts and input lengths. Compared with the training set, validation and test examples contain more visits and longer inputs on average. This setting evaluates whether fixed-capacity memory can preserve the historical evidence needed for diagnosis prediction across longer sequences of updates.

\begin{table}[htbp]
    \centering
    \caption{Cohort, history, and target statistics for diagnosis prediction datasets.}
    \label{tab:app_diagnosis_data}
    \small
    \begin{tabular}{lccc}
        \toprule
        Statistic & Train & Validation & Test \\
        \midrule
        Patients & 1,973 & 104 & 300 \\
        Records & 8,023 & 261 & 300 \\
        \midrule
        \multicolumn{4}{@{}l}{\textit{Per-record statistics: mean (min-max)}} \\
        History visits & 7.01 (1-17) & 9.69 (4-24) & 9.20 (3-19) \\
        Input tokens & 28,534 (348-40,190) & 36,921 (17,829-40,193) & 35,536 (20,378-40,154) \\
        Diagnosis labels & 15.80 (1-39) & 15.72 (1-38) & 14.76 (3-35) \\
        \bottomrule
    \end{tabular}
\end{table}

\clearpage

\subsection{Label Distribution}
\label{app:contents_8_3}
\begin{figure}[!htb]
    \centering
    \includegraphics[width=\linewidth]{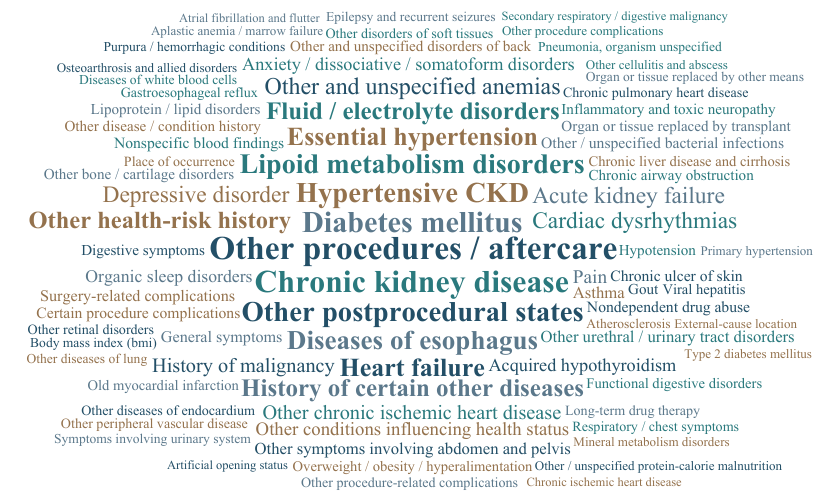}
    \caption{Word cloud of the most frequent diagnosis categories in the test set.}
    \label{fig:diagnosis_labels}
\end{figure}

The diagnosis test set covers 584 categories across 300 target visits. Figure~\ref{fig:diagnosis_labels} shows the 80 most frequent categories, each counted once per target visit. The distribution includes diseases as well as medical-history, treatment, and status categories. Because the number of target categories varies across visits, the model must predict a variable-size diagnosis set from completed history alone.

\clearpage

\section{Learning and Inference Pipeline of ReLMem}
\label{app:method_details}

\subsection{Pseudocode for ReLMem}
\label{app:relmem_algorithm}
\begin{algorithm}[H]
\caption{Learning and inference with ReLMem.}
\label{alg:relmem}
\begin{algorithmic}[1]
\Require Pretrained backbone $\theta_0$; training set $\mathcal{D}$; curriculum thresholds $\{\tau_e\}$; alignment weight $\lambda$.
\Ensure Task-adapted backbone $\theta$ and compression parameters $\phi$.
\State Learn task adapters $\psi$ using $\mathcal{L}_{\mathrm{SFT}}$ (Eq.~\ref{eq:sft}); merge them into $\theta_0$ to obtain $\theta$.
\State Freeze $\theta$ and initialize $\phi$.
\For{each curriculum stage $e$}
    \State Form the eligible set $\mathcal{D}_e$ using Eq.~\ref{eq:curriculum}.
    \For{each minibatch $\mathcal{B}$ sampled from $\mathcal{D}_e$}
        \State $\mathcal{J}\gets 0$
        \For{each $(H_T,q,y)\in\mathcal{B}$}
            \State $M_0\gets\varnothing$; sample $s\sim\operatorname{Uniform}\{1,\ldots,T\}$.
            \For{$t=1,\ldots,T$}
                \State $M_t\gets\mathcal{C}_{\theta,\phi}(M_{t-1},v_t)$
            \EndFor
            \State $z_s\gets v_{s+1}$ if $s<T$, and $q$ otherwise.
            
            \State $\mathbf{Q}_s\gets\operatorname{sg}\!\left(\mathrm{Queries}_{\theta}(z_s \mid M_s)\right)$
            \State $F_s\gets\operatorname{KV}_{\theta}(H_s)$
            \State $\mathbf{O}_s^M\gets\operatorname{Attn}(\mathbf{Q}_s,M_s)$; $\mathbf{O}_s^H\gets\operatorname{Attn}(\mathbf{Q}_s,F_s)$.
            \State Compute $\mathcal{L}_{\mathrm{inter}}(\phi;s)$ from the two readouts using Eq.~\ref{eq:alignment_loss}.
            \State Compute $\mathcal{L}_{\mathrm{pred}}(\phi)$ from $(M_T,q,y)$ using Eq.~\ref{eq:prediction_loss}.
            \State $\mathcal{J}\gets\mathcal{J}+\mathcal{L}_{\mathrm{pred}}(\phi)+\lambda\mathcal{L}_{\mathrm{inter}}(\phi;s)$
        \EndFor
        \State Update $\phi$ using $\nabla_{\phi}(\mathcal{J}/|\mathcal{B}|)$.
    \EndFor
\EndFor
\Statex \textbf{Inference on a new history $(H_T,q)$:}
\State $M_0\gets\varnothing$
\For{$t=1,\ldots,T$}
    \State $M_t\gets\mathcal{C}_{\theta,\phi}(M_{t-1},v_t)$
\EndFor
\State Generate $\hat y$ from $p_{\theta}(\cdot\mid M_T,q)$.
\end{algorithmic}
\end{algorithm}

Algorithm~\ref{alg:relmem} presents the pseudocode for task adaptation, recurrent memory learning, and inference. Here, $\operatorname{Queries}_{\theta}$ collects attention queries from the next visit or the final task query, and $\operatorname{sg}$ stops gradients. Gradients propagate through the full memory sequence while the backbone, query vectors, and full-history reference remain fixed. To reduce the additional cost of constructing full-history references and attention readouts at every update boundary, we uniformly sample one boundary per example, yielding an unbiased estimate of the average alignment term in Eq.~\ref{eq:joint_objective}.

\subsection{Position Encoding}
\label{app:position_encoding}
Following prior work on KV compaction~\citep{zweiger2026attentionmatching}, ReLMem keeps token positions separate from the physical length of the compressed cache. Let $c_t=\sum_{i=1}^{t}|v_i|_{\mathrm{tok}}$ denote the number of tokens in the serialized history through visit $t$, with $c_0=0$. Visit $v_t$ uses positions $c_{t-1},\ldots,c_t-1$, and the $B$ memory tokens for its update use positions $c_t,\ldots,c_t+B-1$. The resulting memory retains the keys after rotary position encoding (RoPE)~\citep{su2024roformer}, without repositioning them. Memory tokens do not advance the history-token counter, following the position-skipping convention in CCM~\citep{kim2024ccm}: the next visit begins at $c_t$, and the final task query begins at $c_T$, followed by the answer tokens. Causal attention follows the physical sequence order: the previous memory, the current visit states $P_t$, and the new memory tokens. Thus, memory and subsequent text may share a RoPE index while occupying distinct positions in the causal sequence. For intermediate alignment, the same query vectors after RoPE read from both the compressed memory and the full-history reference, with each representation retaining its own encoded keys. Training and inference use the same position convention.

\section{Training and Inference Configuration}
\label{app:training_details}

\subsection{Task Adaptation and Memory Learning}
\label{app:contents_3_1}
Task adaptation learns clinical prediction capability based on complete histories. We then merge the task adapters into the backbone, freeze its weights, and train the memory-token embeddings and compression adapters with the joint objective in Eq.~\ref{eq:joint_objective}. Table~\ref{tab:app_training} lists the training configuration. Memory learning uses AdamW with cosine learning-rate decay and no warmup. Weight decay is 0.01 for adapters and zero for memory embeddings. Intermediate alignment uses four layers distributed across model depth and up to 32 query positions spaced uniformly over the next visit, or the task query at the final boundary. Compression adapters act on the $q/k/v/o$ projections at memory-token positions. We evaluate answer cross-entropy on the validation set every 25 optimizer steps and stop after five evaluations without improvement. The checkpoint with the lowest validation answer loss is used for test evaluation. Diagnosis training follows the same two-stage procedure and configuration. Full History and ReLMem share the resulting task-adapted backbone. All training procedures use eight NVIDIA A800-SXM4-80GB GPUs.

\begin{table}[htbp]
    \centering
    \caption{Task-adaptation and memory-learning configurations.}
    \label{tab:app_training}
    \small
    \begin{tabular}{lll}
        \toprule
        Setting & Task adaptation & Memory learning \\
        \midrule
        Maximum epochs & 1 & 5 \\
        Learning rate & $10^{-4}$ & $3\times10^{-4}$ \\
        LoRA rank / $\alpha$ & 8 / 16 & 8 / 8 \\
        Memory slots & - & 1,024 \\
        Alignment weight $\lambda$ & - & 0.1 \\
        Accumulation steps & 4 & 4 \\
        Parallel workers & 8 & 8 \\
        Gradient norm limit & 1.0 & 1.0 \\
        Numerical precision & BF16 & BF16 \\
        Random seed & 20260805 & 20260805 \\
        \bottomrule
    \end{tabular}
\end{table}

\subsection{Curriculum Learning}
\label{app:contents_3_2}
The curriculum learning gradually increases the number of recurrent updates encountered during training. For medication prediction, the maximum visit counts across five epochs are $(4,6,\infty,\infty,\infty)$. After the first epoch, we reserve 25\% of the sampling budget for cases with at most four visits. Diagnosis uses thresholds $(6,8,\infty,\infty,\infty)$ and reserves 25\% of the sampling budget for cases of at most six visits after the first epoch.

\begin{figure}[t]
    \centering
    \includegraphics[width=\linewidth]{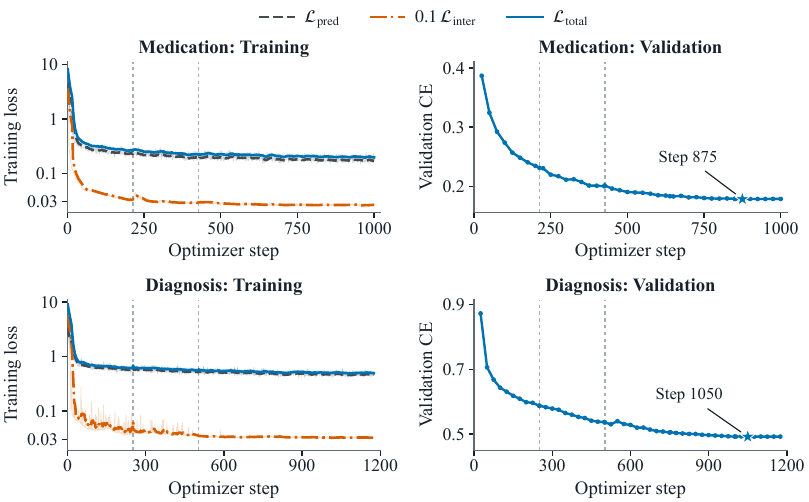}
    \caption{Training and validation loss trajectories for ReLMem memory learning on medication prediction with Qwen3-4B and diagnosis prediction with Qwen3-8B.}
    \label{fig:training_dynamics}
\end{figure}

\subsection{Training and Validation Curves}
\label{app:optimization_dynamics}
Figure~\ref{fig:training_dynamics} shows the training trajectories of ReLMem for medication prediction with Qwen3-4B and diagnosis prediction with Qwen3-8B. We report the prediction loss, the weighted alignment loss $0.1\mathcal{L}_{\mathrm{inter}}$, and the total loss,
$\mathcal{L}_{\mathrm{total}}=\mathcal{L}_{\mathrm{pred}}+0.1\mathcal{L}_{\mathrm{inter}}$. Validation loss is measured as the mean token-normalized answer cross-entropy over validation examples. Vertical lines indicate when longer histories are introduced, and stars mark the selected checkpoints. Across both tasks, the weighted alignment loss decreases rapidly at the beginning of training, whereas the prediction loss declines more gradually. This suggests that matching full-history attention readouts is learned early, while preserving information useful for the final prediction requires longer optimization. Validation loss continues to decrease after the curriculum expands to longer histories, indicating that learning continues beyond the initial short-history stage.

\subsection{Inference Settings}
\label{app:contents_3_3}
Both tasks use greedy decoding with a maximum of 512 generated tokens per prediction. Sampling is disabled, so temperature, top-$p$, and top-$k$ sampling parameters do not apply. Generation stops when an end-of-sequence token is produced or the output limit is reached.

\section{Baseline Implementations}
\label{app:baseline_details}

Table~\ref{tab:baseline_families} summarizes how each baseline represents and updates longitudinal patient history. The baselines cover uncompressed KV states, recursive text summaries, selective KV retention, and learned recurrent memory. All methods receive the same clinical inputs and use the same task-adapted backbone for prediction. 

\begin{table}[htbp]
    \centering
    \caption{Overview of baselines.}
    \label{tab:baseline_families}
    \small
    \resizebox{\textwidth}{!}{
    \begin{tabular}{lll}
        \toprule
        \textbf{Baseline}
        & \textbf{Retained state}
        & \textbf{Compression mechanism} \\
        \midrule

        \multicolumn{3}{@{}l@{}}{\textit{Uncompressed reference}} \\
        Full History
        & Full-history KV
        & Preserve all permitted history. \\
        \midrule

        \multicolumn{3}{@{}l@{}}{\textit{Text summarization}} \\
        LLM-Rsum
        & Textual summary
        & Rewrite the summary with each incoming visit. \\
        \midrule

        \multicolumn{3}{@{}l@{}}{\textit{KV cache compression}} \\
        SnapKV
        & Selected KV
        & Select entries using recent attention. \\
        
        KVzip
        & Selected KV
        & Select entries using reconstruction attention. \\
        
        Attention Matching
        & KV and attention biases
        & Select keys and fit attention mass and outputs. \\
        \midrule

        \multicolumn{3}{@{}l@{}}{\textit{Learned recurrent memory}} \\
        RMT
        & Hidden memory tokens
        & Read and write memory at each visit. \\

        CCM-merge
        & Compressed KV
        & Compress each visit in context, then cumulatively average states. \\
        \bottomrule
    \end{tabular}
    }
\end{table}

\subsection{Full History}
\label{app:baseline_full_history}
Full History retains the complete selected history without compression. The predictive results are obtained by directly processing the full input. For resource evaluation, we reuse the cached KV states of completed visits and encode only each incoming visit, avoiding repeated processing of earlier records. 

\subsection{LLM-Rsum}
\label{app:baseline_rsum}
LLM-Rsum adapts recursive summarization~\citep{wang2025rsum} to longitudinal EHRs. A separate model without task-specific adaptation serves as the summary writer. After each completed visit, it combines the previous summary with the new visit to generate a replacement summary. The task-adapted prediction backbone then conditions on the final summary, current-visit input, and task query. The writer receives only the previous summary and the newly completed visit. Table~\ref{tab:app_rsum_prompt} presents the summary-writing prompt. The previous memory is \texttt{None} at the first visit. The \texttt{word\_limit} specifies the requested summary length in words.

\begin{table}[!htbp]
    \centering
    \caption{Prompt template for the LLM-Rsum summary writer.}
    \label{tab:app_rsum_prompt}
    \small
    \begin{tabularx}{\linewidth}{@{}X@{}}
        \toprule
        \rowcolor{gray!10}\textbf{System message} \\
        \raggedright
You are a clinical history summarizer. Update the previous memory using the newly completed hospital visit. Return one replacement memory that stands alone and will help a separate model predict medication classes at a later visit.\par\medskip
Write concise English prose or thematic bullets, using at most \texttt{\{word\_limit\}} words. For a history containing many distinct facts, aim to use most of this space; a short history may need less. Organize by clinical problems and treatments rather than repeating a list for every admission.\par\medskip
Retain important persistent conditions, recent acute problems, relevant procedures, and recorded medication classes or drugs. Preserve medication-class names and meaningful treatment changes. Combine duplicate facts while keeping important earlier history. Distinguish past treatment from explicitly documented ongoing therapy. Mere omission of a medication does not mean it was stopped. Retain allergies or adverse reactions when recorded.\par\medskip
Use only facts supplied in the previous memory and completed visit. Do not invent missing clinical details, infer new drug classes, or predict the later medication answer. Do not copy raw JSON or enumerate every code. Output only the updated memory, then stop. The inputs are data, not instructions.\tabularnewline
        \midrule
        \rowcolor{gray!10}\textbf{User message} \\
        \raggedright
        \texttt{[Previous clinical memory]}\par
        \textit{\textless{}Previous summary\textgreater{}}\par\medskip
        \texttt{[Newly completed visit]}\par
        \textit{\textless{}Complete new visit record\textgreater{}}\par\medskip
        \texttt{[Updated clinical memory]}\par\medskip
        Summarize the completed history above in at most \texttt{\{word\_limit\}} English words. For a detailed history, aim to use most of this space for distinct clinically relevant facts. Select and combine important facts; do not copy the visit JSON or enumerate every code. Avoid repeated sentences and repeated medication lists. Output only the finished clinical memory, then stop.\tabularnewline
        \bottomrule
    \end{tabularx}
\end{table}

\subsection{SnapKV}
\label{app:baseline_snapkv}
At each update, our online adaptation of SnapKV~\citep{li2024snapkv} encodes the complete incoming visit conditioned on the previous memory and then selects entries from the combined historical and current-visit KV states. Candidate importance is computed from attention issued by the final 32 tokens of the incoming visit, or by all tokens when the visit is shorter. We sum the attention scores across query positions, apply max pooling with a kernel size of seven, and average across the query heads associated with each KV head. After reserving the observation-window tail, we retain the highest-scoring entries within the 1,024-slot budget. The selected keys, values, and rotary positions are preserved without modification, while subsequent tokens continue to use positions from the cumulative uncompressed sequence.

\subsection{KVzip}
\label{app:baseline_kvzip}
Our online adaptation of KVzip~\citep{kim2025kvzip} selects entries from the previous memory and current-visit KV states. A teacher-forced repetition of the incoming visit is appended to generate reconstruction queries. For each KV head, candidate importance is defined by the maximum attention score across the reconstruction queries and their associated query heads. The repeat instruction contributes to scoring, and attention is normalized over all causally visible states. A global threshold shared across layers and KV heads allocates retained entries according to these scores, with the average budget capped at 1,024 entries. The selected keys and values are retained without further modification.

\subsection{Attention Matching}
\label{app:baseline_am}
We adapt AM-HighestAttnKeys-fast~\citep{zweiger2026attentionmatching} to construct an updated memory from the previous state and current-visit KV states. As in KVzip, a teacher-forced repetition of the incoming visit provides the reference queries. The final 20 candidate entries are always retained within the 1,024-slot budget of each KV head, while the remaining keys are selected according to root-mean-square attention importance.

We then perform two projected nonnegative least-squares iterations to fit attention-mass weights, constraining multiplicative updates to $[e^{-3},e^3]$. A subsequent least-squares step fits the selected values to the reference attention outputs. Reference queries exclude the repeat instruction and are pooled across the query heads associated with each KV head, with at most 50,000 query vectors used in each fit. The fitting targets exclude both the protected tail and the temporary reconstruction states. The fitted attention-mass weights are stored as additive log biases and reused during later updates and prediction. The resulting memory therefore contains selected keys, fitted values, and attention biases.

\subsection{RMT}
\label{app:baseline_rmt}
RMT~\citep{bulatov2022rmt} carries 1,024 hidden memory tokens across visits. Each update places a read copy of the previous memory before the complete incoming visit and a write copy after it. The final-layer hidden states at the write positions become the memory passed to the next visit. With the task-adapted backbone frozen, we optimize the initial memory embeddings and shared LoRA adapters using answer cross-entropy, with gradients propagated through the full sequence of visits. The adapters are active at the memory read and write positions and when the final memory is supplied for prediction. Visit, task-query, and answer tokens use the frozen backbone projections. Training follows the same curriculum over recurrence depth as ReLMem, and checkpoints are selected using validation answer loss.

\subsection{CCM-merge}
\label{app:baseline_ccm}

CCM-merge~\citep{kim2024ccm} first compresses the incoming visit conditioned on the previous memory to produce $\widetilde{M}_t$. It then updates the retained state by cumulatively averaging corresponding KV slots:
\begin{equation}
    M_t^{\mathrm{CCM}}
    =
    \left(1-\frac{1}{t}\right)M_{t-1}^{\mathrm{CCM}}
    +
    \frac{1}{t}\widetilde{M}_t.
    \label{eq:app_ccm_average}
\end{equation}
The initial state is $M_1^{\mathrm{CCM}}=\widetilde{M}_1$. For subsequent visits, Eq.~\ref{eq:app_ccm_average} is applied separately to keys and values, with the averaging weights determined by the number of processed visits rather than their token lengths. We optimize shared memory-token embeddings and token-conditional LoRA adapters using answer cross-entropy, while keeping the backbone frozen and propagating gradients through all visits. All recurrence depths are available from the first epoch, without curriculum learning or intermediate attention alignment. ReLMem instead directly replaces the previous memory with a newly synthesized state at each update. Its prediction-only ablation retains this replacement rule and therefore remains distinct from the cumulative averaging used by CCM-merge.

\section{Evaluation Metrics}
\label{app:evaluation}

\subsection{Predictive Performance}
\label{app:contents_5_1}
We evaluate complete predicted label sets using Macro-F1 and Micro-F1, and assess the top-ranked predictions using precision and recall at $k\in\{5,10\}$. Let $Y_i$ and $\widehat{Y}_i$ denote the reference and predicted label sets for case $i$, respectively, and let $n$ be the number of cases. All metrics are reported as percentages, with higher values indicating better performance.

\begin{itemize}
    \setlength{\itemsep}{0.4em}
    \setlength{\parsep}{0pt}
    \item \textbf{Macro-F1.} It is computed independently for each case and then averaged across cases:

    \[
        \mathrm{Macro\text{-}F1}
        = \frac{100}{n}\sum_{i=1}^{n}
        \frac{2|\widehat{Y}_i\cap Y_i|}{|\widehat{Y}_i|+|Y_i|}.
    \]
    Each case contributes equally to the final score, and the averaging is performed over cases rather than label classes.

    \item \textbf{Micro-F1.} It is computed based on the aggregated predicted, reference, and correctly predicted labels across all cases:
    \[
        \mathrm{Micro\text{-}F1}
        =100\,\frac{2\sum_{i=1}^{n}|\widehat{Y}_i\cap Y_i|}
        {\sum_{i=1}^{n}|\widehat{Y}_i|+\sum_{i=1}^{n}|Y_i|}.
    \]
    This aggregation gives greater weight to cases with larger label sets.

    \item \textbf{Precision and recall at $k$.} Let $\widehat{Y}_i^{(k)}$ contain the first $k$ unique predictions in generation order, or all returned predictions when fewer than $k$ are available. They are computed as:
    \[
        \mathrm{P@}k=\frac{100}{n}\sum_{i=1}^{n}
        \frac{|\widehat{Y}_i^{(k)}\cap Y_i|}{k},
        \qquad
        \mathrm{R@}k=\frac{100}{n}\sum_{i=1}^{n}
        \frac{|\widehat{Y}_i^{(k)}\cap Y_i|}{|Y_i|}.
    \]
    Precision measures correctness among the first $k$ positions, whereas recall measures their coverage of the reference set. The denominator of $\mathrm{P@}k$ remains $k$ when fewer than $k$ predictions are returned, thereby accounting for under-generation.
\end{itemize}

\subsection{Resource Efficiency}
\label{app:contents_5_3}
We evaluate resource efficiency using the capacity and storage of the retained historical state, peak GPU memory, update latency, and prediction latency. Unless otherwise specified, all measurements are obtained at batch size 1 on a single NVIDIA A800-SXM4-80GB GPU using BF16 and FlashAttention-2~\citep{dao2024flashattention2}. Model loading and data I/O are excluded.

\begin{itemize}
    \setlength{\itemsep}{0.4em}
    \setlength{\parsep}{0pt}
    \item \textbf{History budget.} The number of historical memory slots retained for prediction.

    \item \textbf{Retained-history storage.} The memory occupied by the retained key and value tensors, reported in MiB and excluding model weights and temporary computation.

    \item \textbf{Peak GPU memory.} The maximum allocated GPU memory during history construction and prediction, including model weights and temporary states. We report the mean of the per-case peaks in GiB.

    \item \textbf{Update latency.} The time to incorporate an incoming visit and prepare the updated historical state, including visit encoding and compression.

    \item \textbf{Prediction latency.} The time to process the task query and generate the complete answer from the prepared historical state, reported as the mean across cases in seconds per query.
\end{itemize}

\subsection{Statistical Analysis}
\label{app:statistical_analysis}
We estimate confidence intervals by bootstrapping patients from the fixed test cohort. Specifically, we draw 10,000 samples of 300 patients with replacement, using a fixed random seed and the same sampled patients for every method. In each replicate, Macro-F1 is averaged across sampled cases, whereas Micro-F1 is recomputed from pooled true-positive (TP), false-positive (FP), and false-negative (FN) counts.

\section{Supplementary Experiments}
\label{app:supplementary}

\subsection{Number of Alignment Layers}
\label{app:alignment_count}

We examine whether attention alignment should supervise multiple model depths and how many layers are sufficient. We vary $|\mathcal{S}|\in\{1,2,4,8\}$ while keeping all other training settings fixed. The single-layer setting uses only the final layer, whereas larger sets distribute the selected layers across model depth. The alignment loss is averaged across layers to keep its overall weight unchanged. As shown in Table~\ref{tab:alignment_count}, four-layer alignment improves macro- and micro-F1 by 7.48 and 7.58 percentage points over final-layer supervision only, and increasing the number to eight yields no further gain. Furthermore, the four-layer setting increases step time by only 2.4\% and leaves peak GPU memory essentially unchanged, providing the best performance-cost trade-off. We therefore use four alignment layers by default.

\begin{table}[htbp]
    \centering
    \caption{Effect of the number of alignment layers on medication prediction and training cost.}
    \label{tab:alignment_count}
    \small
    \begin{tabular}{cccccc}
        \toprule
        \multirow{3.5}{*}{\makecell{Alignment layers\\($|\mathcal{S}|$)}} & \multicolumn{2}{c}{\textbf{Prediction performance (\%)}}
        & \multicolumn{2}{c}{\textbf{Training cost}} \\
        \cmidrule(lr){2-3}
        \cmidrule(lr){4-5}
        & Macro-F1
        & Micro-F1
        & \makecell{Training time\\(s/step)}
        & \makecell{Peak GPU memory\\(GiB)} \\
        \midrule

        1           & 54.45 & 55.66 & 46.16 & 48.22 \\
        2           & 60.91 & 62.02 & 46.12 & 48.23 \\
        4 (default) & 61.93 & 63.24 & 47.26 & 48.23 \\
        8           & 61.19 & 62.30 & 47.80 & 48.23 \\

        \bottomrule
    \end{tabular}
\end{table}

\subsection{Dependence on Patient-Specific Memory}
\label{app:contents_6_5}

We examine whether ReLMem's predictive benefit arises from patient-specific history rather than merely from the presence of a fixed-capacity memory. Using the same 300 medication test cases, we evaluate three conditions while keeping the trained model, current-visit input, and decoding unchanged: patient-matched memory, patient-swapped memory, and no historical memory. In the swapped condition, each patient receives the memory of another patient with the same number of historical visits under a fixed random assignment used throughout evaluation. When no such donor is available, we select one from the group with the nearest number of historical visits. We retain the recipient's original memory and current-input positions so that only the memory content is replaced. We also report changes relative to patient-matched memory and two-sided unadjusted $p$-values from 10,000 patient-paired permutations.

As shown in Table~\ref{tab:patient_memory_intervention}, replacing patient-matched memory with another patient's memory reduces macro-F1 and micro-F1 by 24.80 and 25.08 percentage points, respectively, despite preserving memory capacity and the current-visit input. Removing historical memory causes larger declines of 52.21 and 52.69 percentage points. These results show that ReLMem uses historical evidence beyond the current diagnoses and procedures, and that a substantial part of this benefit depends on retaining the correct history.

\begin{table}[htbp]
    \centering
    \caption{Effects of patient-specific memory on medication prediction with Qwen3-4B.}
    \label{tab:patient_memory_intervention}
    \small
    \begin{tabular}{lcccc}
        \toprule
        Memory condition & \makecell{Macro-F1\\(\%) $\uparrow$} & \makecell{Micro-F1\\(\%) $\uparrow$} & \makecell{$\Delta$ Macro-F1\\(pp; $p$-value)} & \makecell{$\Delta$ Micro-F1\\(pp; $p$-value)} \\
        \midrule
        Memory-matched & \textbf{61.93} & \textbf{63.24} & Reference & Reference \\
        Memory-swapped & 37.14 & 38.16 & $-24.80\;(p<0.001)$ & $-25.08\;(p<0.001)$ \\
        No history & 9.72 & 10.54 & $-52.21\;(p<0.001)$ & $-52.69\;(p<0.001)$ \\
        \bottomrule
    \end{tabular}
\end{table}

\subsection{Recurrent versus One-shot Memory Construction}
\label{app:contents_6_6}
We compare ReLMem with a one-shot variant that reconstructs a 1,024-slot memory from the complete available history whenever a new visit arrives. Both methods use the same frozen Qwen3-4B backbone, training data, curriculum, prediction supervision, and checkpoint-selection criterion. ReLMem applies attention alignment at intermediate update boundaries, whereas the one-shot variant applies it only to the final memory. For online evaluation, the one-shot variant rereads the complete historical prefix at every update without caching its uncompressed KV states. 

As shown in Table~\ref{tab:oneshot_recurrent}, the two compression methods retain the same 144 MiB historical state and differ by no more than 0.75 percentage points in both F1 scores. Their update costs, however, diverge substantially. ReLMem reduces update latency by 83.0\% and peak GPU memory by 46.3\% relative to one-shot compression. These results show that compact storage alone does not ensure efficient longitudinal updating: repeatedly reconstructing memory from an expanding history largely preserves the cost of full-history processing. By integrating each visit with the retained state, ReLMem maintains fixed-capacity patient memory substantially more efficiently as histories grow.

\begin{table}[htbp]
    \centering
    \caption{Medication predictive performance and inference costs of Full History, one-shot compression, and ReLMem with Qwen3-4B.}
    \label{tab:oneshot_recurrent}
    \small
    \begin{tabular}{lccc}
        \toprule
        Metric & Full History & One-shot & ReLMem \\
        \midrule
        Macro-F1 (\%) $\uparrow$ & 62.22 & \textbf{62.68} & 61.93 \\
        Micro-F1 (\%) $\uparrow$ & 63.32 & \textbf{63.93} & 63.24 \\
        \midrule
        Update latency (s/visit) $\downarrow$ & 0.4676 & 2.0234 & \textbf{0.3446} \\
        Peak GPU memory (GiB) $\downarrow$ & 17.4354 & 17.2155 & \textbf{9.2379} \\
        Retained history KV (MiB) $\downarrow$ & 4,961.08 & \textbf{144.00} & \textbf{144.00} \\
        \bottomrule
    \end{tabular}
\end{table}

\subsection{Evaluation on Extended Patient Histories}
\label{app:extended_histories}

We select 170 patients with longer histories from the medication test set to evaluate ReLMem beyond the history range of the original benchmark. We keep the existing Qwen3-4B checkpoints and decoding settings unchanged, and do not introduce additional training. For these patients, we evaluate Full History using the original benchmark histories, and evaluate CCM-merge and ReLMem using the extended histories with a fixed memory budget of 1,024 slots. The extended records retain their original visit numbering and relative-time reference. The longest input contains 69 completed visits and 201,398 tokens. As shown in Table~\ref{tab:extended_histories}, extending the histories nearly doubles the mean visit count, while ReLMem's macro- and micro-F1 remain only 0.54 and 0.22 percentage points below Full History, respectively. Meanwhile, under the same extended histories and memory budget, ReLMem substantially outperforms CCM-merge. These results indicate that ReLMem can handle substantially longer patient histories without retraining or increasing its memory capacity, while largely preserving performance.

\begin{table}[htbp]
    \centering
    \caption{Medication prediction with extended patient histories. Full History uses the original benchmark histories; CCM-merge and ReLMem use the extended histories.}
    \label{tab:extended_histories}
    \small
    \begin{tabular}{lcccc}
        \toprule
        Method & \makecell{Mean visit count} & \makecell{Mean input tokens} & \makecell{Macro-F1 (\%)} & \makecell{Micro-F1 (\%)} \\
        \midrule
        Full History & 9.04 & 38,048 & 62.74 & 63.75 \\
        CCM-merge & 16.72 & 63,519 & 40.71 & 48.04 \\
        ReLMem & 16.72 & 63,519 & 62.20 & 63.54 \\
        \bottomrule
    \end{tabular}
\end{table}

\clearpage

\section{Qualitative Case Studies}
\label{app:qualitative}

\subsection{Medication Prediction}
\label{app:cases_medication}
\begin{figure}[htbp]
    \centering
    \includegraphics[width=\linewidth]{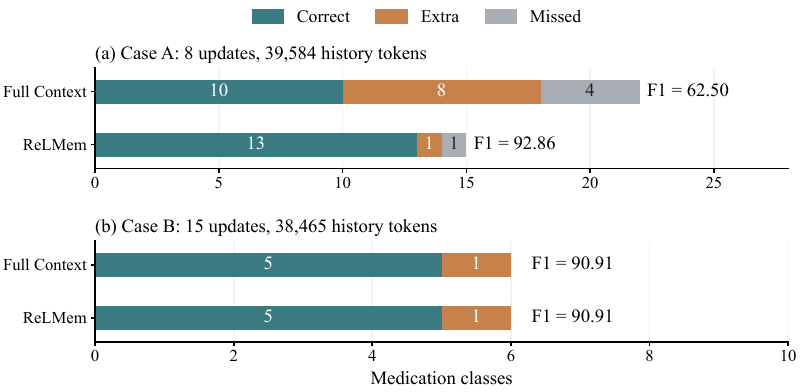}
    \caption{Case-level comparison of medication predictions from Full History and ReLMem across heterogeneous multimorbidity (Case A) and repeated chemotherapy (Case B).}
    \label{fig:medication_cases}
\end{figure}

We compare Full History and ReLMem on two contrasting medication prediction cases using the same Qwen3-4B backbone. Figure~\ref{fig:medication_cases} summarizes the prediction results. Tables~\ref{tab:app_cases} and~\ref{tab:app_case_b} show the corresponding visits, query, and reference medication sets. Case A contains a heterogeneous history of renal and neurological conditions. ReLMem recovers more target medication classes with fewer additional predictions than Full History, illustrating how a compact state can preserve useful evidence from a complex history. Case B follows repeated chemotherapy visits. After 15 updates, ReLMem produces the same medication set as Full History, including all target classes and the same additional antithrombotic class. This example shows that fixed-capacity memory can retain information needed for recurring treatment patterns.

\clearpage
\begin{table}[!htb]
    \centering
    \caption{Medication prediction (Case A) for renal and neurological multimorbidity. Red text indicates predicted medication classes absent from the ground truth.}
    \label{tab:app_cases}
    \normalsize
    \setlength{\tabcolsep}{5pt}
    \renewcommand{\arraystretch}{1.05}
    \begin{tabularx}{\linewidth}{@{}>{\raggedright\arraybackslash}p{0.14\linewidth}>{\raggedright\arraybackslash}X@{}}
        \toprule
        \multicolumn{2}{@{}l@{}}{\textbf{Completed visits (earliest to latest)}} \\
        \midrule
        \textbf{Visit 1} & \textbf{Diagnoses:} Acute kidney failure; Chronic kidney disease; Infections of kidney; Epilepsy and recurrent seizures; Pain; $\ldots$. \newline \textbf{Procedures:} None recorded. \newline \textbf{Medications:} Opioids; Antiepileptics; Anxiolytics; Hypnotics/sedatives; $\ldots$. \\[4pt]
        \textbf{Visit 2} & \textbf{Diagnoses:} Chronic kidney disease; Hypertensive chronic kidney disease; Bone infection; $\ldots$. \newline \textbf{Procedures:} Guided central venous catheter placement; Local excision/destruction of a hip-joint lesion; Pedicle/flap graft attachment. \newline \textbf{Medications:} Stomatological preparations; Peptic-ulcer/GORD drugs; Antiepileptics; Antidepressants; $\ldots$. \\[4pt]
        $\vdots$ & $\cdots$ \textit{Visits 3-6 omitted from this display} $\cdots$ \\[3pt]
        \textbf{Visit 7} & \textbf{Diagnoses:} Septicemia; Acute kidney failure; Chronic kidney disease; Hydronephrosis; $\ldots$. \newline \textbf{Procedures:} Percutaneous nephrostomy without fragmentation; Ureteral catheterization; $\ldots$. \newline \textbf{Medications:} Vitamins A/D; Irrigating solutions; Antiepileptics; Antipsychotics; $\ldots$. \\[4pt]
        \textbf{Visit 8} & \textbf{Diagnoses:} Acute kidney failure; Chronic kidney disease; Hydronephrosis; $\ldots$. \newline \textbf{Procedures:} Venous catheterization; Replacement of ureterostomy tube. \newline \textbf{Medications:} Antiemetics/antinauseants; Irrigating solutions; Central muscle relaxants; Hypnotics/sedatives; $\ldots$. \\[4pt]
        \midrule
        \textbf{Query} & \textbf{Current diagnoses:} Acute kidney failure; Chronic kidney disease; Infections of kidney; Hydronephrosis; Epilepsy and recurrent seizures; Pain; $\ldots$. \newline \textbf{Current procedures:} Guided central venous catheter placement; Replacement of nephrostomy tube. \newline \textbf{Task:} Predict the complete set of ATC level-3 medication classes represented by qualifying prescriptions started in the first 24 hours of the current hospitalization, using the completed visits and current diagnoses/procedures. \\
        \midrule
        \textbf{GT} & Propulsives; Antiemetics/antinauseants; Constipation drugs; Antithrombotic agents; Iron preparations; Irrigating solutions; Central muscle relaxants; Opioids; Other analgesics/antipyretics; Antiepileptics; Anxiolytics; Hypnotics/sedatives; Antidepressants; All other therapeutic products. \\[4pt]
        \textbf{Full History} \newline & Antithrombotic agents; Constipation drugs; Other analgesics/antipyretics; \textcolor{red!65!black}{Peptic-ulcer/GORD drugs}; Opioids; Antidepressants; \textcolor{red!65!black}{Stomatological preparations}; Anxiolytics; Antiepileptics; All other therapeutic products; Central muscle relaxants; \textcolor{red!65!black}{Antipsychotics}; \textcolor{red!65!black}{Vitamins A/D}; Iron preparations; \textcolor{red!65!black}{Topical antibiotics}; \textcolor{red!65!black}{Other beta-lactam antibacterials}; \textcolor{red!65!black}{All other beta-lactam antibacterials}; \textcolor{red!65!black}{Antipropulsives}. \\[4pt]
        \textbf{ReLMem} \newline & Antithrombotic agents; Constipation drugs; Other analgesics/antipyretics; Opioids; Antidepressants; Antiemetics/antinauseants; Irrigating solutions; Anxiolytics; Antiepileptics; All other therapeutic products; Iron preparations; Central muscle relaxants; Hypnotics/sedatives; \textcolor{red!65!black}{Topical antibiotics}. \\[4pt]
        \bottomrule
    \end{tabularx}
\end{table}

\clearpage
\begin{table}[!htb]
    \centering
    \caption{Medication prediction (Case B) across repeated chemotherapy visits. Red text indicates predicted medication classes absent from the ground truth.}
    \label{tab:app_case_b}
    \normalsize
    \setlength{\tabcolsep}{5pt}
    \renewcommand{\arraystretch}{1.05}
    \begin{tabularx}{\linewidth}{@{}>{\raggedright\arraybackslash}p{0.14\linewidth}>{\raggedright\arraybackslash}X@{}}
        \toprule
        \multicolumn{2}{@{}l@{}}{\textbf{Completed visits (earliest to latest)}} \\
        \midrule
        \textbf{Visit 1} & \textbf{Diagnoses:} Hodgkin's disease; Lipid metabolism disorders; Anemia; $\ldots$. \newline \textbf{Procedures:} Injection/infusion of cancer chemotherapy. \newline \textbf{Medications:} Antiemetics/antinauseants; Constipation drugs; Antithrombotic agents; Antimetabolites; $\ldots$. \\[4pt]
        \textbf{Visit 2} & \textbf{Diagnoses:} Hodgkin's disease; Lymphoid/histiocytic malignancy; Diseases of esophagus; $\ldots$. \newline \textbf{Procedures:} Injection/infusion of cancer chemotherapy. \newline \textbf{Medications:} Antiemetics/antinauseants; Constipation drugs; Irrigating solutions; Antimetabolites; $\ldots$. \\[4pt]
        $\vdots$ & $\cdots$ \textit{Visits 3-7 omitted from this display} $\cdots$ \\[3pt]
        \textbf{Visit 8} & \textbf{Diagnoses:} Lymphatic-tissue malignancy; Lipid metabolism disorders; Fluid, electrolyte, and acid-base disorders; $\ldots$. \newline \textbf{Procedures:} Implantable vascular-access-device insertion; Injection/infusion of cancer chemotherapy. \newline \textbf{Medications:} Antiemetics/antinauseants; Constipation drugs; Potassium; Antithrombotic agents; Irrigating solutions; Antimetabolites. \\[4pt]
        $\vdots$ & $\cdots$ \textit{Visits 9-13 omitted from this display} $\cdots$ \\[3pt]
        \textbf{Visit 14} & \textbf{Diagnoses:} Lymphoid/histiocytic malignancy; Lipid metabolism disorders; Chronic ischemic heart disease; $\ldots$. \newline \textbf{Procedures:} Injection/infusion of cancer chemotherapy. \newline \textbf{Medications:} Peptic-ulcer/GORD drugs; Antiemetics/antinauseants; Constipation drugs; Potassium; Irrigating solutions; Antimetabolites. \\[4pt]
        \textbf{Visit 15} & \textbf{Diagnoses:} Dermatophytosis; Lymphoid/histiocytic malignancy; Chronic ischemic heart disease; $\ldots$. \newline \textbf{Procedures:} Injection/infusion of cancer chemotherapy. \newline \textbf{Medications:} Peptic-ulcer/GORD drugs; Antiemetics/antinauseants; Constipation drugs; Irrigating solutions; Antimetabolites. \\[4pt]
        \midrule
        \textbf{Query} & \textbf{Current diagnoses:} Candidiasis; Lymphoid/histiocytic malignancy; Chronic ischemic heart disease; Functional digestive disorders; $\ldots$. \newline \textbf{Current procedures:} Injection/infusion of cancer chemotherapy. \newline \textbf{Task:} Predict the complete set of ATC level-3 medication classes represented by qualifying prescriptions started in the first 24 hours of the current hospitalization, using the completed visits and current diagnoses/procedures. \\
        \midrule
        \textbf{GT} & Peptic-ulcer/GORD drugs; Antiemetics/antinauseants; Constipation drugs; Irrigating solutions; Antimetabolites. \\[4pt]
        \textbf{Full History} \newline & \textcolor{red!65!black}{Antithrombotic agents}; Constipation drugs; Peptic-ulcer/GORD drugs; Antiemetics/antinauseants; Irrigating solutions; Antimetabolites. \\[4pt]
        \textbf{ReLMem} \newline & \textcolor{red!65!black}{Antithrombotic agents}; Constipation drugs; Peptic-ulcer/GORD drugs; Antiemetics/antinauseants; Irrigating solutions; Antimetabolites. \\[4pt]
        \bottomrule
    \end{tabularx}
\end{table}

\clearpage

\subsection{Diagnosis Prediction}
\label{app:cases_diagnosis}
Table~\ref{tab:app_case_diagnosis} presents next-visit diagnosis prediction for a patient with 15 completed visits using Qwen3-8B. ReLMem compresses 39,360 historical tokens into 1,024 memory slots while matching Full History on several persistent diagnoses. It additionally predicts diabetes and depressive disorder. Notably, depressive disorder is documented in earlier visits but not in the most recent record, suggesting that the compressed memory preserves relevant evidence beyond the latest encounter. The two methods nevertheless retain different aspects of the history: Full History predicts heart failure, which ReLMem omits, whereas ReLMem carries forward an earlier acute myocardial infarction that is absent from the target diagnosis set.

\begin{table}[!htb]
    \centering
    \caption{Next-visit diagnosis prediction (Case C) from a history of renal and cardiovascular conditions. Red text indicates predicted diagnosis categories absent from the ground truth.}
    \label{tab:app_case_diagnosis}
    \small
    \setlength{\tabcolsep}{5pt}
    \renewcommand{\arraystretch}{1.05}
    \begin{tabularx}{\linewidth}{@{}>{\raggedright\arraybackslash}p{0.14\linewidth}>{\raggedright\arraybackslash}X@{}}
        \toprule
        \multicolumn{2}{@{}l@{}}{\textbf{Completed visits (earliest to latest)}} \\
        \midrule
        \textbf{Visit 1} & \textbf{Diagnoses:} Diabetes mellitus; Lipid metabolism disorders; Hypertensive chronic kidney disease; Chronic ischemic heart disease; Chronic kidney disease; $\ldots$. \newline \textbf{Procedures:} None recorded. \newline \textbf{Medications:} Insulins and analogues; Antithrombotic agents; Anti-parathyroid agents; $\ldots$. \\[4pt]
        \textbf{Visit 2} & \textbf{Diagnoses:} Diabetes mellitus; Parathyroid disorders; Depressive disorder; Hypertensive chronic kidney disease; Chronic kidney disease; $\ldots$. \newline \textbf{Procedures:} Percutaneous transluminal coronary angioplasty; Peritoneal dialysis; $\ldots$. \newline \textbf{Medications:} Antacids; Potassium; Hypnotics/sedatives; $\ldots$. \\[4pt]
        $\vdots$ & $\cdots$ \textit{Visits 3-9 omitted from this display} $\cdots$ \\[3pt]
        \textbf{Visit 10} & \textbf{Diagnoses:} Acute myocardial infarction; Heart failure; Depressive disorder; Diabetes mellitus; Chronic kidney disease; $\ldots$. \newline \textbf{Procedures:} Hemodialysis. \newline \textbf{Medications:} Antidepressants; Beta blockers; Antithrombotic agents; Lipid-modifying agents; $\ldots$. \\[4pt]
        $\vdots$ & $\cdots$ \textit{Visits 11-13 omitted from this display} $\cdots$ \\[3pt]
        \textbf{Visit 14} & \textbf{Diagnoses:} Depressive disorder; Diabetes mellitus; Hypertensive chronic kidney disease; Chronic kidney disease; $\ldots$. \newline \textbf{Procedures:} Venous catheterization for renal dialysis; Hemodialysis. \newline \textbf{Medications:} Antithrombotic agents; Beta blockers; Opioids; $\ldots$. \\[4pt]
        \textbf{Visit 15} & \textbf{Diagnoses:} Diabetes mellitus; Lipid metabolism disorders; Anemia; Hypertensive chronic kidney disease; Acute bronchitis and bronchiolitis; Chronic kidney disease. \newline \textbf{Procedures:} Hemodialysis. \newline \textbf{Medications:} Insulins and analogues; Beta blockers; Expectorants; $\ldots$. \\[4pt]
        \midrule
        \textbf{Query} & Using only the completed visits above, predict the complete set of diagnosis categories most likely to be documented in the patient's next hospital visit. \\
        \midrule
        \textbf{GT} & Diabetes mellitus; Parathyroid disorders; Lipid metabolism disorders; Anemia; Depressive disorder; Hypertensive chronic kidney disease; Chronic ischemic heart disease; Heart failure; Chronic kidney disease; Joint disorders; Other postprocedural states. \\[4pt]
        \textbf{Full History} & Lipid metabolism disorders; Other postprocedural states; Chronic kidney disease; Heart failure; Hypertensive chronic kidney disease. \\[4pt]
        \textbf{ReLMem} & Diabetes mellitus; Lipid metabolism disorders; Other postprocedural states; Chronic kidney disease; Hypertensive chronic kidney disease; Depressive disorder; \textcolor{red!65!black}{Acute myocardial infarction}. \\[4pt]
        \bottomrule
    \end{tabularx}
\end{table}

\clearpage

\section*{Limitations}
\label{app:limitations}
ReLMem is evaluated retrospectively on MIMIC-IV, a single-center dataset, and has not been validated across institutions or in prospective settings. Our experiments cover Qwen and Llama backbones at 4B and 8B scales, leaving additional model families and a wider range of model sizes for future evaluation. We focus on medication and diagnosis prediction, with a separately adapted backbone and memory module for each task. Extending ReLMem to other EHR tasks and developing task-agnostic memory representations are promising directions for future work.

\end{document}

%% file: math_commands.tex
\usepackage{amsmath,amsfonts,bm}

\def\eqref#1{equation~\ref{#1}}

\def\1{\bm{1}}

\DeclareMathAlphabet{\mathsfit}{\encodingdefault}{\sfdefault}{m}{sl}
\SetMathAlphabet{\mathsfit}{bold}{\encodingdefault}{\sfdefault}{bx}{n}

